\documentclass[conference]{IEEEtran}
\IEEEoverridecommandlockouts

\usepackage{amsmath,amssymb,amsfonts}
\usepackage{algorithmic}
\usepackage{graphicx}
\usepackage{textcomp}
\usepackage{xcolor}
\usepackage{dblfloatfix}

\usepackage{url}
\usepackage{caption}
\usepackage[square,numbers,sort&compress]{natbib}

\usepackage[caption=false,font=footnotesize]{subfig}
\usepackage{floatrow}

\definecolor{royalpurple}{RGB}{188,66,245}

\definecolor{pptgreen}{RGB}{0,176,80}

\usepackage{setspace}
\AtBeginDocument{%
  \addtolength\abovedisplayskip{-0.3\baselineskip}%
  \addtolength\belowdisplayskip{-0.3\baselineskip}%
  \addtolength\abovedisplayshortskip{-0.3\baselineskip}%
  \addtolength\belowdisplayshortskip{-0.3\baselineskip}%
}

\makeatletter
\newcommand{\linebreakand}{%
  \end{@IEEEauthorhalign}
  \hfill\mbox{}\par
  \mbox{}\hfill\begin{@IEEEauthorhalign}
}
\makeatother

\def\BibTeX{{\rm B\kern-.05em{\sc i\kern-.025em b}\kern-.08em
    T\kern-.1667em\lower.7ex\hbox{E}\kern-.125emX}}
\begin{document}


\title{\LARGE Evaluation of Deep Neural Operator Models toward Ocean Forecasting}
\vskip -0.9truecm

\vskip -0.9truecm
\author{\IEEEauthorblockN{Ellery Rajagopal\textsuperscript{$b$},
Anantha N.\,S.\ Babu\textsuperscript{$a$},
Tony Ryu\textsuperscript{$a$}, 
Patrick J.\ Haley, Jr.\textsuperscript{$a$},
Chris Mirabito\textsuperscript{$a$},
Pierre F.\,J.\ Lermusiaux\textsuperscript{$a, \dagger$}
}
\IEEEauthorblockA{\textit{\textsuperscript{$a$}
Department of Mechanical Engineering, Center for Computational Science and Engineering,}}
\IEEEauthorblockA{\textit{\textsuperscript{$b$}
Department of Electrical Engineering and Computer Science,}}
\IEEEauthorblockA{\textit{
Massachusetts Institute of Technology, Cambridge, MA}}
\IEEEauthorblockA{\textsuperscript{$\dagger$}Corresponding author: pierrel@mit.edu}}

\maketitle
\thispagestyle{plain}
\pagestyle{plain}

\begin{abstract}
    Data-driven, deep-learning modeling frameworks have been recently developed for forecasting time series data. Such machine learning models may be useful in multiple domains including the atmospheric and oceanic ones, and in general, the larger fluids community. The present work investigates the possible effectiveness of such deep neural operator models for reproducing and predicting classic fluid flows and simulations of realistic ocean dynamics. We first briefly evaluate
    the capabilities of such deep neural operator models when trained on a simulated two-dimensional fluid flow past a cylinder. 
    We then investigate their application 
    to forecasting ocean surface circulation in the Middle Atlantic Bight and Massachusetts Bay, learning from high-resolution data-assimilative simulations employed for real sea experiments. 
    We confirm that trained deep neural operator models are capable of predicting idealized periodic eddy shedding. For realistic ocean surface flows and our preliminary study, they can predict several of the features and show some skill, providing potential for future research and applications.
\end{abstract}


\section{Introduction}

In recent years, the fluid dynamics communities including Numerical Weather Prediction (NWP) have been introduced to a collection of data-driven, deep-learning models. These models make use of machine learning architectures such as neural networks, neural operators, and transformers to learn the mapping between one continuous function space to another, and the resulting mapping allows forecasting time series data. 
%
In general, neural operators directly learn the mapping between two infinite-dimensional function spaces. DeepONet \cite{lu2021learning} 
was the first neural operator to learn such mapping combining sub-networks of a branch net and a trunk net. Various modifications and extensions of DeepONet have been proposed to further improve its accuracy and efficiency \cite{lu2022comprehensive, goswami2022physics, kontolati2023learning}. Neural operators were also developed based on graph kernel networks with kernel integration done by message passing using graph networks \cite{li2020neural}. However, these operators are unstable for very deep neural networks. To overcome this, another neural operator was proposed based on the Fourier transform \cite{li2020fourier}. More recently, neural operators have been developed using the wavelet transform \cite{tripura2022wavelet, gupta2021multiwavelet}.
Since their introduction, vision transformers have shown superior performance over other architectures in various image recognition and generation tasks \cite{dosovitskiy2020image}. For high-resolution applications, transformers with self-attention mechanisms can capture fine-scale features with higher accuracy and efficiency than conventional architectures such as convolutional neural networks (CNNs) \cite{guibas2021adaptive}.
%

Two recent examples of models that combine neural nets, neural operators, and transformers are FourCastNet \cite{pathak2022fourcastnet} and Latent DeepONet \cite{kontolati2023learning}. 
FourCastNet combines neural operators, specifically Fourier Neural Operators (FNOs) \cite{li2020fourier}, with Vision Transformers (ViT) \cite{dosovitskiy2020image} to represent the operator that evolves a physical quantity forward in time by one unit of time. 
Latent DeepONet
learns the operator that maps a physical quantity at an initial state to the same quantity evolved forward in time after a (predefined) number of timesteps. It does so by combining the output of two neural nets: one for encoding the input at a fixed number of points and the other for encoding the locations for the output. 



While many of these models have been trained for and applied to forecasting atmospheric variables such as air temperature and wind velocity \cite{pathak2022fourcastnet,bonev2023spherical,tripura2022wavelet,kontolati2023learning}, they do not restrict inputs only to atmospheric quantities. 
That is, one may treat these models as a ``black box" and train them with appropriate tuning of the network architecture such that they take the state of a dynamical system as input and return the state evolved forward in time as output \cite{lu2022comprehensive}. 
This allows applying these 
models to other fields such as ocean or fluid dynamics by choosing the adequate inputs and outputs during training. 


Several studies focus on these models because they offer reduced computational costs during inference \cite{michalowska2023neural} and can be trained directly with diverse types of observational data \cite{lu2022comprehensive}. Many deep neural operator models 
so far aim to reproduce or forecast the system state 
without the aid of governing equations \cite{li2020fourier, bonev2023spherical}. They also avoid the use of adjoint equations for training  \cite{li2020fourier}
since neural operators are differentiable and hence back-propagation does not require adjoint methods as those used in neural ODEs or DDEs  \cite{gupta_lermusiaux_SR2023,gupta_lermusiaux_PRSA2021}.  

In this work, to establish a benchmark, we first apply deep neural operator models to a classic idealized fluid flow, the two-dimensional (2D) flow past a cylinder \cite{zdravkovich1997,kundu2004}. 
We then investigate their application to surface ocean circulation forecasting \cite{haley_lermusiaux_OD2010,haley_et_al_OM2015}. Specifically, we consider the Middle Atlantic Bight \cite{mseas_exercises_awacs_2006,lin_et_al_JOE2010,lermusiaux_et_al_BBN_Oceans2020}
and Massachusetts Bay \cite{lermusiaux_JMS2001,haley_et_al_Oceans2020},
and utilize training and verification data from high-resolution data-assimilative simulations developed for real sea experiments. 

\section{Deep Neural Operator Models}

One is often interested in representing the time evolution of a physical quantity defined over a continuous domain. More precisely, a common goal is to learn the mapping between the quantity at one time to the quantity at another time. 
Despite the continuous analytical representations, one
discretizes the domain for the purpose of computation. 
When applied to discretized problems, the performances of neural networks were shown to vary heavily depending on the choice of discretization \cite{li2020neural}. 
As the ideal discretization for a given problem or class of problems is rarely known, standard neural networks are commonly limited by the specifics of the discretization. 

As opposed to learning a mapping between two finite-dimensional vector spaces, neural operators aim to learn a mapping between two infinite-dimensional function spaces \cite{kovachki2023neural}. A key advantage over standard neural networks is then that neural operators are discretization invariant, thus avoiding the fundamental issue encountered by standard networks. 
One is no longer restricted by the specific choice of discretization to learn the time evolution of the desired quantities. 


\subsection{DeepONet}

DeepONet was the first
neural operator to learn such a mapping, taking advantage of the universal approximation theorem of operators that suggests a 
sufficiently large neural network with a single hidden layer can accurately approximate any nonlinear continuous operator up to a small approximation error \cite{chen1995universal,lu2021learning}. As a result, DeepONet can be used to learn the operator which maps a physical quantity at an initial state to the same quantity evolved forward in time after a given number of timesteps. It does so by combining the output of the trunk net, which encodes the input at a fixed number of points, with the output of the branch net, which encodes the locations for the output. The trunk-branch structure is directly inspired by the universal approximation theorem of operators. 

Consider an input function $u$ and an operator $G$, we are interested in the output $(G(u))(y)$ where $y$ is a spatial location in the domain of $G(u)$. DeepONet combines a branch network and a trunk network to obtain the output as follows \cite{lu2021learning}:
\begin{equation}
    (G(u))(y)\approx\sum_{k=1}^p b_k(u) \otimes t_k(y) + b_0  \; 
\end{equation}
where $p$ is the number of branch networks, $b_k$ the output of each branch network, 
$t_k$ the output of the trunk network,
and $b_0$ the bias. 
The specific architecture of the sub-networks could be designed according to the desired application \cite{lu2022comprehensive}.
Latent DeepONet \cite{kontolati2023learning} is an extension of DeepONet to learn the operator from the latent representations of the input and output spaces. This is implemented utilizing a multi-layer autoencoder which is pre-trained to encode the inputs and decode the outputs of DeepONet.


In fluid dynamics, DeepONet has been used to study linear instabilities in high-speed boundary layers and their inverse problem \cite{di2023neural, hao2023instability}. 
It also inferred multiscale bubble growth and dynamics \cite{lin2021seamless}. Latent DeepONet had successes in fluid convection and in simulating shallow water equations \cite{kontolati2023learning}.

\subsection{Kernel-Based Neural Operator Models}

Neural operator mappings were also learned through the use of graph kernel networks \cite{li2020neural}, which rely on an underlying quadrature-like or Nystr{\"o}m approximation formulation \cite{nystrom1930praktische} linking multiple grid points to one set of parameters for the network. A key difference between neural networks and these neural operators is the inclusion of kernel integration in the formulation. With $\vec{v}_0$ serving as the original input, the output $\vec{v}_N$ of a standard neural network is defined as the $N$-th iteration of the recursion \cite{li2020neural},
\begin{equation}
    \vec{v}_{i+1} = \sigma(W_i\vec{v}_i + b_i)
\end{equation}
In this formulation, $W$ is a linear transformation, $b$ the bias, and $\sigma$ the nonlinear activation function. 
The output of a graph kernel network, on the other hand, follows a similar but different recursion, switching the vector $\vec{v}_i$ for the function $v_i(x)$ where $x$ is any spatial location in the domain of interest. 
With $v_0(x)$ serving as the original input, the output $v_N(x)$ is defined as the $N$-th iteration of the recursion
\begin{equation}
    v_{i+1}(x) = \sigma\left(W_iv_i(x) + (K(a;\Phi)v_i)(x)\right)
\end{equation}
where $W$ is again a linear transformation and $\sigma$ a nonlinear activation function. $K(a;\Phi)$ is the kernel integral transformation parameterized by $\Phi$ and defined as
\begin{equation}
    (K(a;\Phi)v_i)(x) = \int_D \kappa_{\Phi}(x, y, a(x), a(y);\phi)v_i(y)dy \; 
\end{equation}
where $\kappa_{\Phi}$ is a neural network function with parameters $\Phi$, $a$ is a function of space, and $x,y$ are spatial locations in the domain of interest. 

%

The now well-known FNO architecture was developed by expanding upon the idea of graph kernel networks and learning parameters directly in Fourier space through the use of what is known as Fourier layers. These layers are discretization-invariant because they can learn from and evaluate functions that are discretized arbitrarily \cite{li2020neural}. The FNO architecture imposes that the kernel function $\kappa_{\Phi}$ not depend on the function $a$, along with $\kappa_{\Phi}(x, y) = \kappa_{\Phi}(x - y)$. This turns the kernel integration operator into a convolution operator, allowing for the use of the FFT for a boost in computational speed \cite{li2020fourier}. 

In fluid dynamics applications, FNOs have been used for predicting turbulent flows \cite{peng2022attention, li2023long}. Their accuracy in predicting cylindrical wakes for a range of subcritical Reynolds numbers has been studied \cite{renn2023forecasting}. FNOs have also been used in modeling multi-fidelity and multi-phase flows \cite{lyu2023multi, wen2022u}.

FNOs were extended to Adaptive FNOs (AFNOs) that make use of the self-attention mechanism of the ViT architecture to address the shortcomings of using images as input for FNOs \cite{guibas2021adaptive}. 
This improves the scalability and robustness of the resulting predictions by imposing a block diagonal structure on the weight tensor, sharing weights adaptively across tokens, and sparsifying tokens using soft-thresholding and shrinkage operations. 
The AFNO model is the backbone of FourCastNet, which was originally designed to predict global weather patterns using atmospheric state variables such as surface wind speed, precipitation, and atmospheric water vapor \cite{pathak2022fourcastnet}. For this work, we investigate the preliminary use of FourCastNet and Latent DeepONet within an oceanographic context.

\subsection{Model Inputs and Outputs, and Training}

While both FourCastNet (FCN) and DeepONet (L-DoN) are used for operator learning, the structure of the operators for both models is different. For a collection of images $\{x_1, \dots, x_n\}$, FCN learns the operator which maps $x_i \mapsto x_{i+1} \ \forall i \in \{1, \dots, n-1\}$. That is, we have 
\begin{equation}
    x_{i+1} = FCN(x_i) \ \forall i \in \{1, \dots, n-1\}.
\end{equation}
Each training point for FCN is thus the pair 
\begin{equation}
    (x_i, x_{i+1}) \ \forall \ i \in \{1, \dots, n-1\}.
\end{equation}

For the same set of images, L-DoN learns the mapping from $x_1 \mapsto \{x_1, \dots, x_N\}$. That is, it maps the first image to all N images, allowing for the prediction of $N$ images in ``one shot". This means that
\begin{equation}
    \{x_1, \dots, x_N\} = DON(x_1).
\end{equation}

To train FCN, we first partition our training data into predefined batches of equal size, with each batch containing contiguous subsets of the training time series. We then loop through our collection of batches and perform a forward pass of the model with each batch. We evaluate the loss function on the output of each forward pass. 
Using PyTorch's automatic differentiation, we use stochastic gradient descents to adjust the model weights. This process constitutes one epoch of training, and we train for a predefined number of epochs.
With L-DoN, input values are mapped to many data values. Inputs are selected such that the union of their outputs covers the training set. The training is otherwise similar to that of FCN.

%

\section{Flow Past a Cylinder}

We first briefly evaluate the properties and accuracy of the two deep neural operator architectures when trained on the 2D velocity of a fluid flowing past a cylinder (FpC) or idealized island. 
The key parameter is the non-dimensional Reynolds number, 
$\rm{Re}=V_{\infty}L/\nu$ \cite{kundu2004}, where $V_{\infty}$ is the fluid velocity at infinity from the obstacle, $L$ the projected width of the obstacle, and $\nu$ the dynamic viscosity of the fluid.  For $\rm{Re}$ within approximately 5 and 40, the flow downstream of the obstacle is symmetric.  For $\rm{Re}$ greater than 40, the flows exhibit periodic, asymmetric vortex shedding known as von K\'arm\'an vortex streets. For $\rm{Re}$ increasing beyond 200, more complex aperiodic patterns appear, up to fully turbulent flows.

To create the data sets for these
idealized flow cases, we employ our MIT-MSEAS 2.29 finite-volume (FV) framework \cite{lermusiaux_2.29_notes,ueckermann_and_lermusiaux_MSEAS2012}. It contains a finite volume solver for 2D nonhydrostatic fluid and ocean dynamics. 
It has been very useful for diverse ocean dynamics process studies and for the incubation of advanced schemes and methodologies.

\subsection{Test Case Description}

%
Many evaluations were completed, but we only show results for a single set-up. 
The domain is a rectangular channel with a height of 3~m and a width of 20~m. The inlet velocity is set to $2\;\textrm{m}/\textrm{s}$. The diameter of the cylinder is set at 1~m, meaning with our kinematic viscosity of $0.01\;\textrm{m}^2/\textrm{s}$, the $\rm{Re}$ is~200. 

The MIT-MSEAS 2.29 numerical simulation uses 50×100 second-order finite volumes and a time-step of 0.005~s. 
The diffusion operator is discretized using central boundary fluxes, a total variation diminishing (TVD) scheme with a monotonized central (MC) symmetric flux limiter is used for advection, and a rotational incremental pressure correction scheme is used for time integration. 
We simulate the flow for a total of 500~s.
With the above parameters, we observe fully developed
periodic, asymmetric vortex shedding after 50~s.
An eddy sheds approximately every 2~s, so we find a shedding period of about 4~s. 
For the machine learning evaluation, 
the horizontal and vertical velocities, $u$ and $v$, respectively, are sub-sampled every 0.01~s and saved for the whole $50\times100$ numerical domain, but only starting at time $t=50\;\textrm{s}$ after the flow has developed. 
That is, these 450 seconds provide 45000 snapshots of dimension $50\times100\times2$ (each snapshot is a $50\times100$ array for each of $u$~and~$v$ in m/s). 

%

\subsection{Learning Results}

We find that both FCN and L-DoN are capable of predicting the present idealized FpC flows with periodic, asymmetric vortex shedding. They can do this for long periods of time compared to the shedding frequency, but the phase and details of the structures can be lost over time. Varying the training procedures, network parameters, and hyperparameters can significantly alter the performance. 

For good FCN and L-DoN performance, we trained on the first 25,000 snapshots, validated on the next 10,000, and predicted on the remaining 10,000. 
For FCN, we selected a global batch size of 2, size of image patches of~$2\times2$, number of AFNO layers (depth) of 3, and embedded channel size of 96. The loss function employed was the L2-norm. Training was executed for 150 to 200 epochs.
For L-DoN, we utilize similar hyperparameters as in \cite{kontolati2023learning} but with latent dimension of 16, batch size of 2, and size of the final hidden layer ($p$) of 4. 
The loss function employed was the L2-norm. Training was executed for 1000 epochs. 

\begin{figure*}
    \centering
    \subfloat[]{\includegraphics[width=0.20\textwidth]{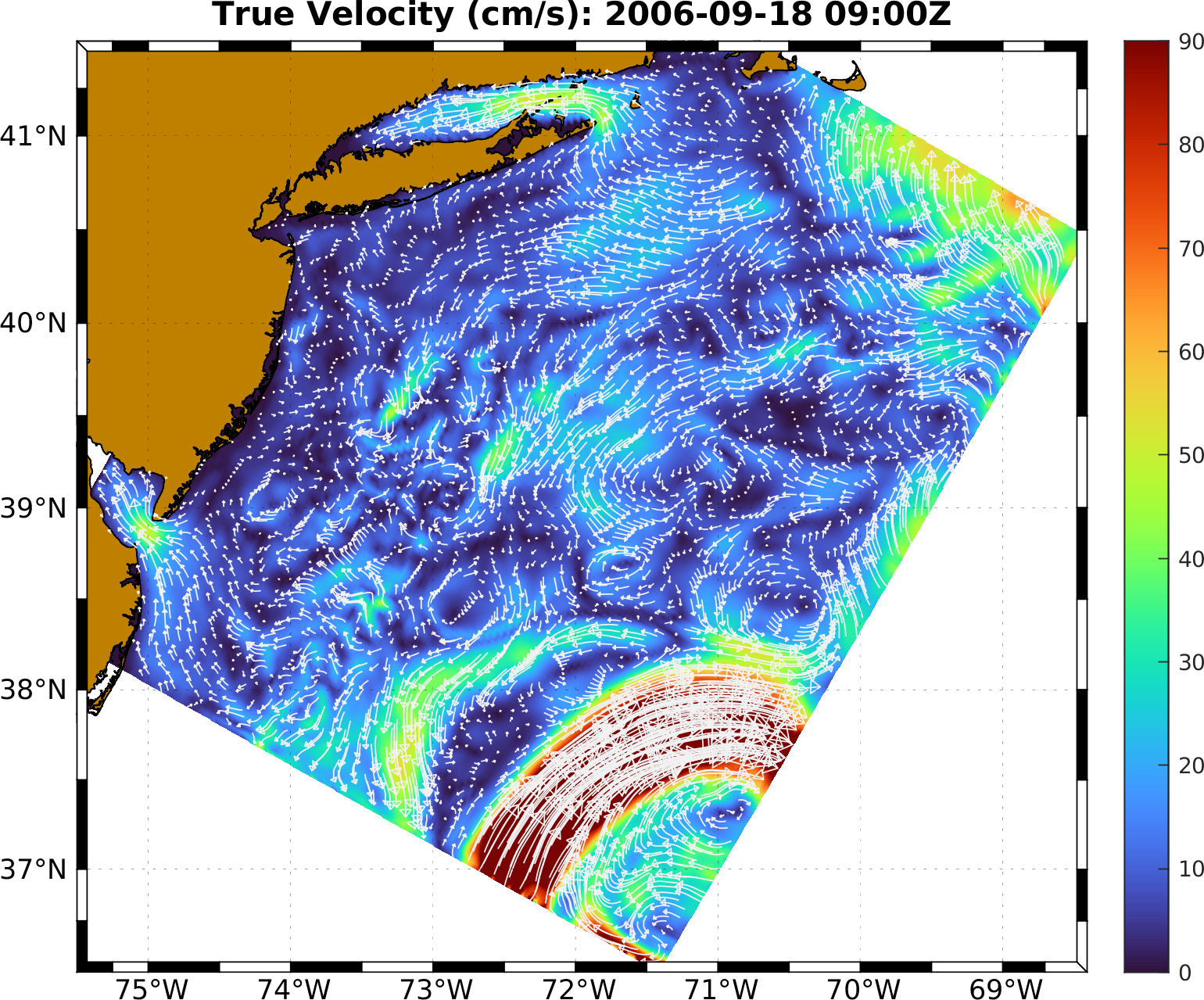}}
    \subfloat[]{\includegraphics[width=0.20\textwidth]{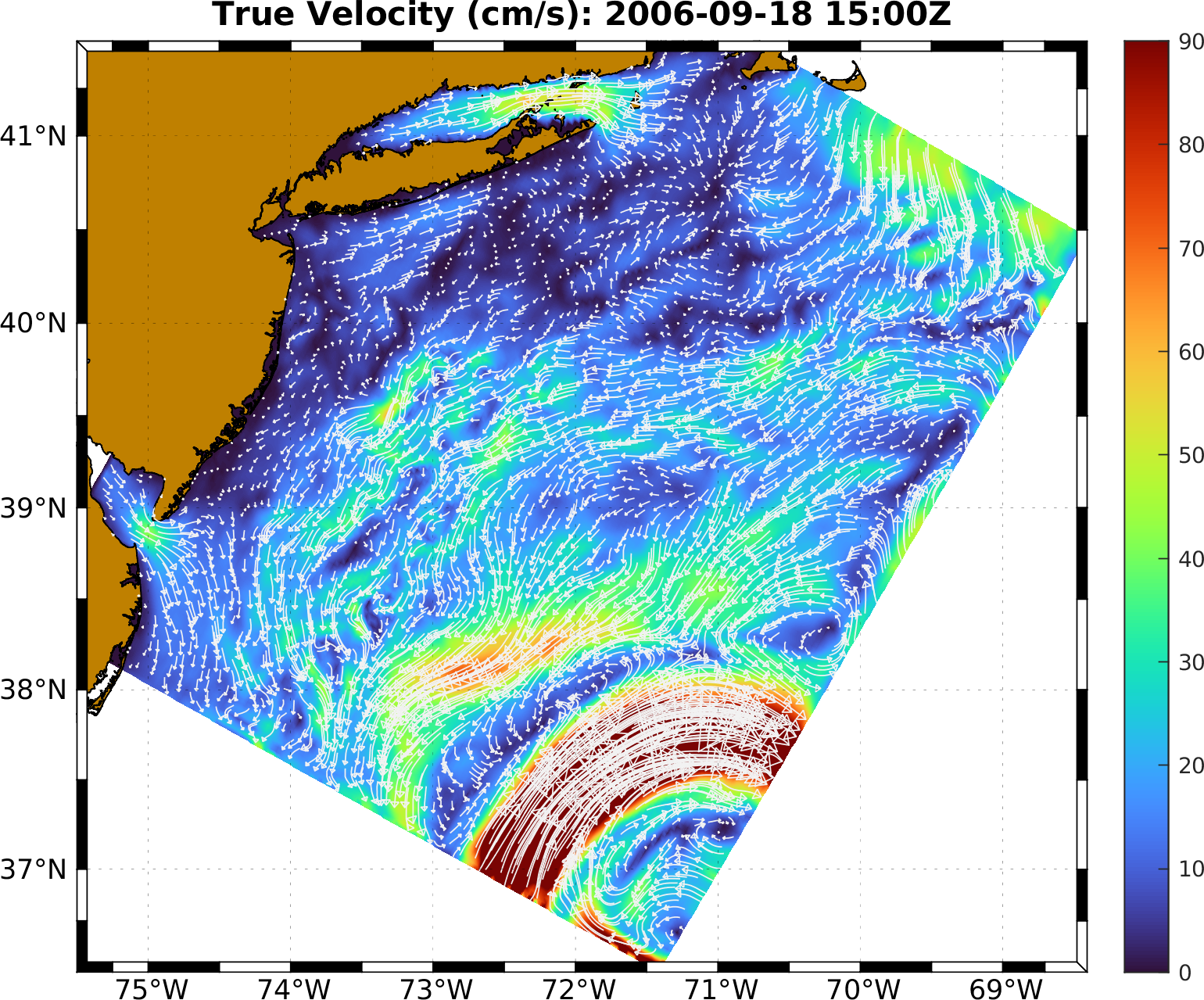}}
    \subfloat[]{\includegraphics[width=0.20\textwidth]{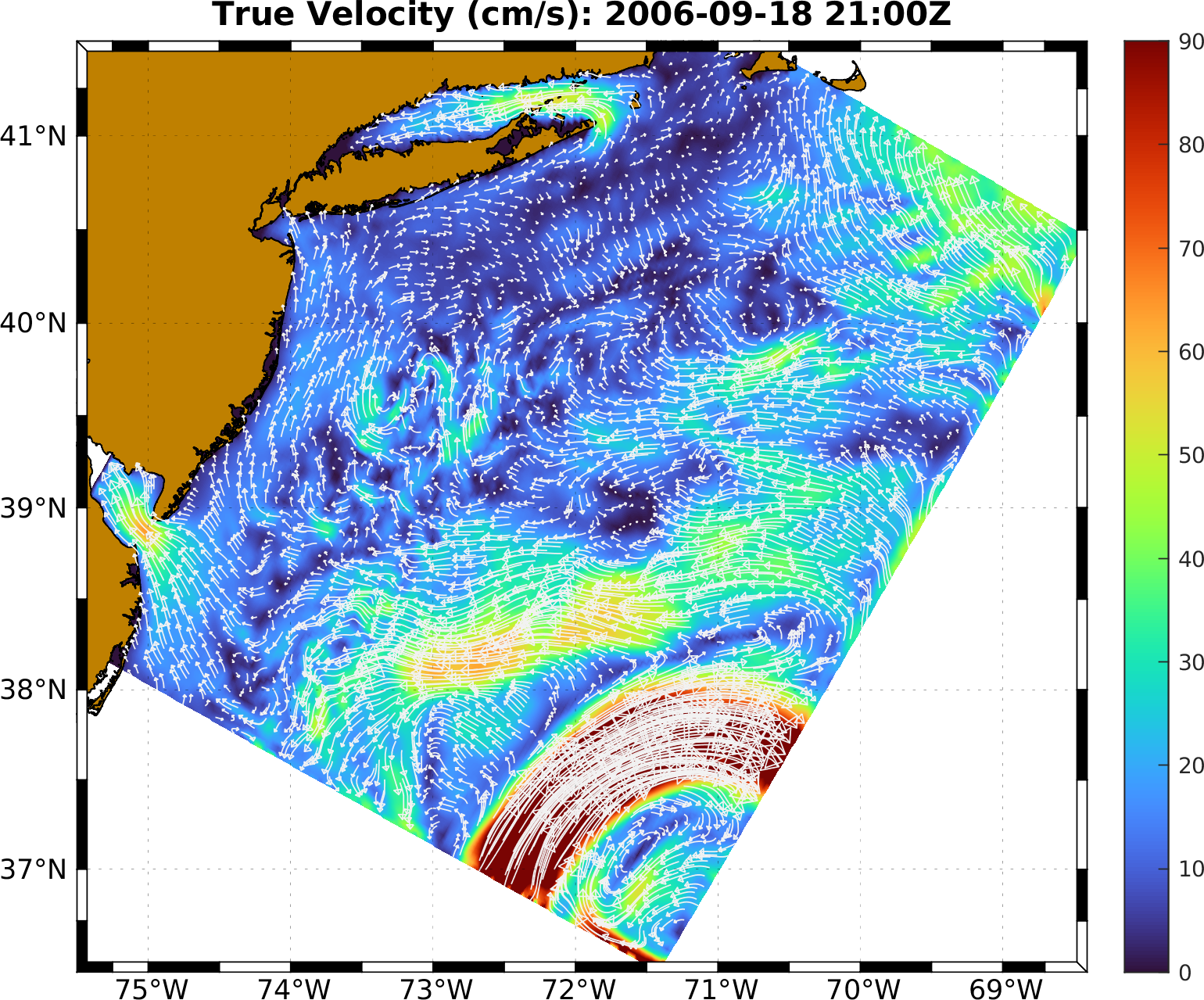}}
    \subfloat[]{\includegraphics[width=0.20\textwidth]{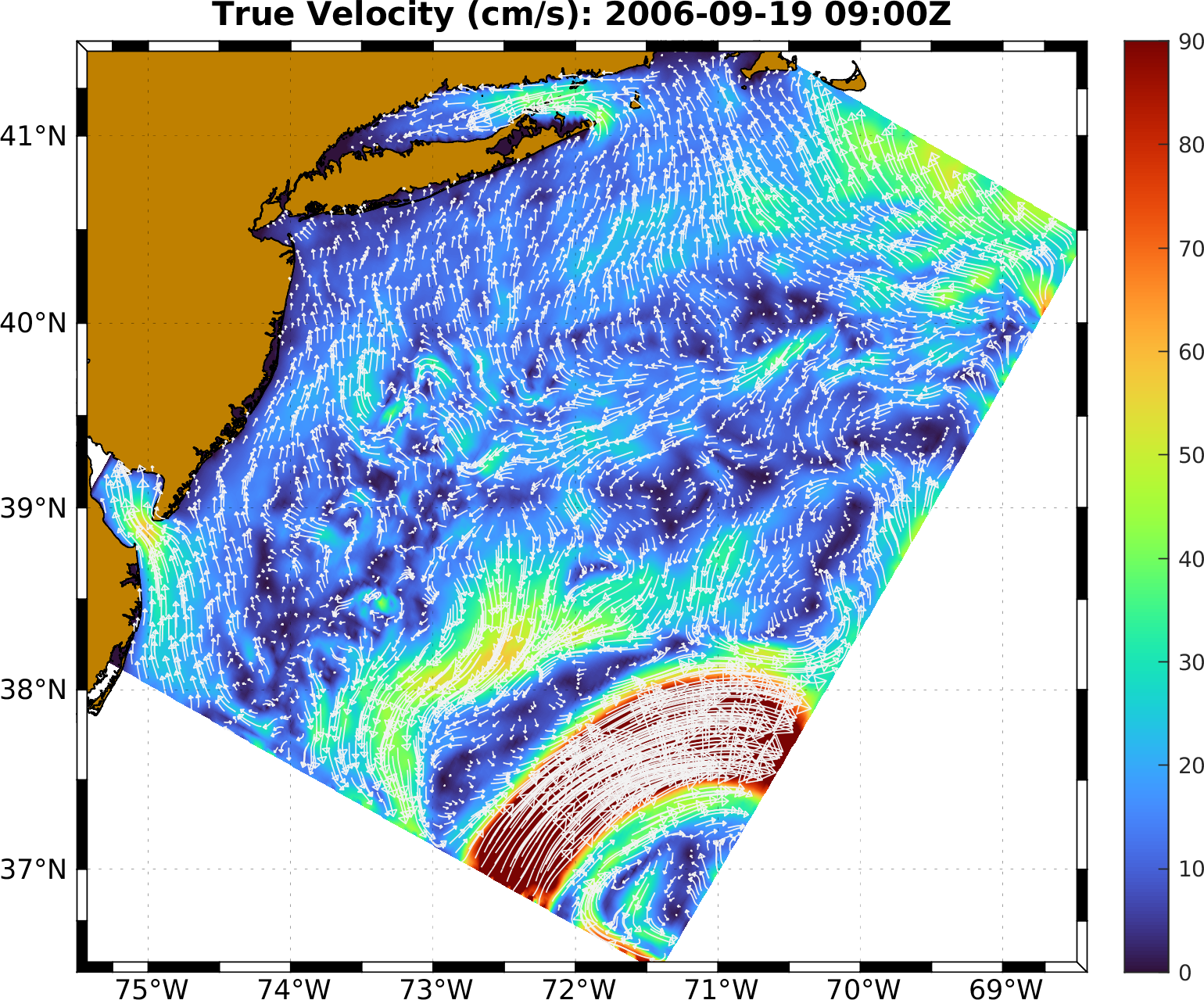}}
    \subfloat[]{\includegraphics[width=0.20\textwidth]{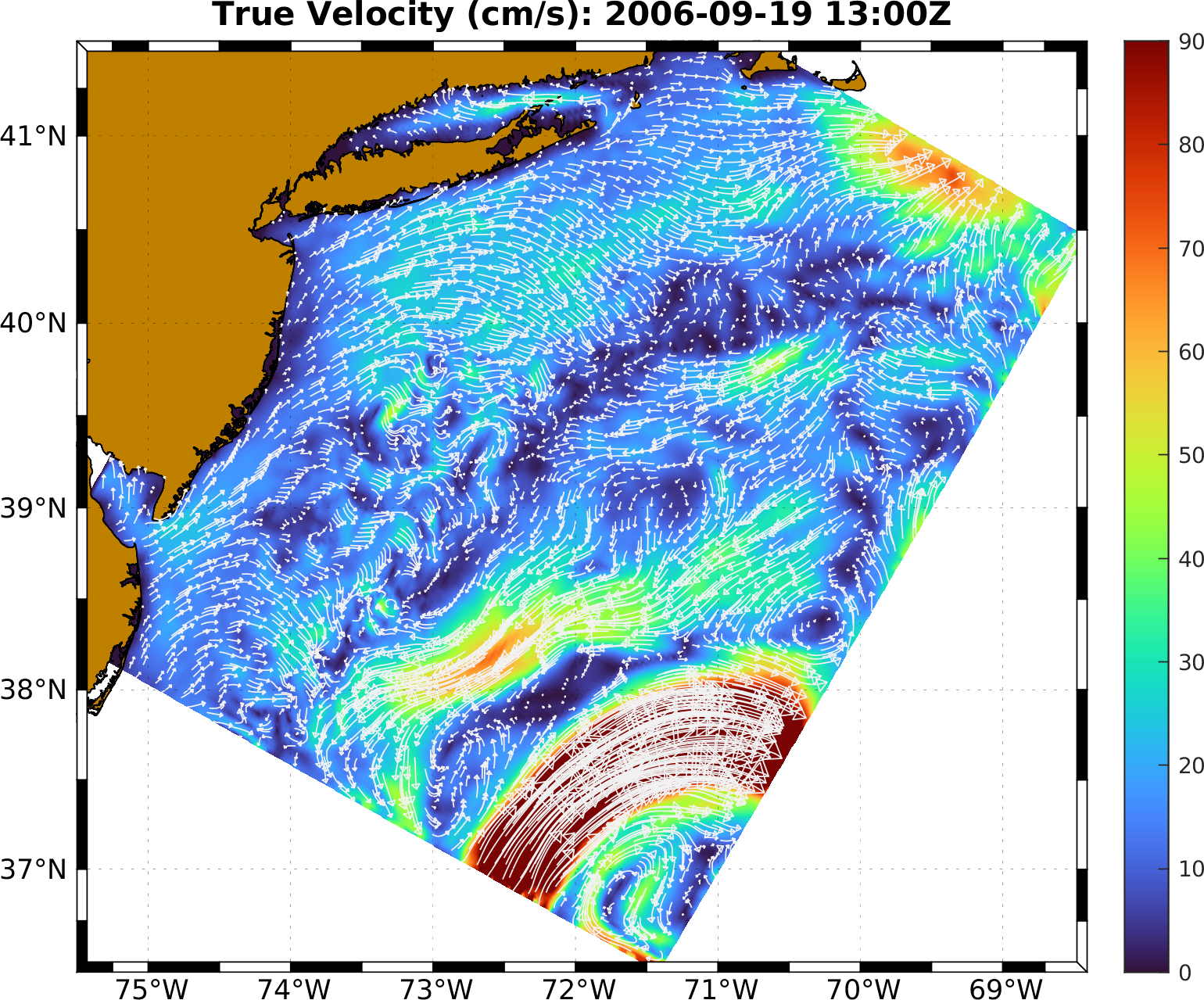}} \\
    \subfloat[]{\includegraphics[width=0.20\textwidth]{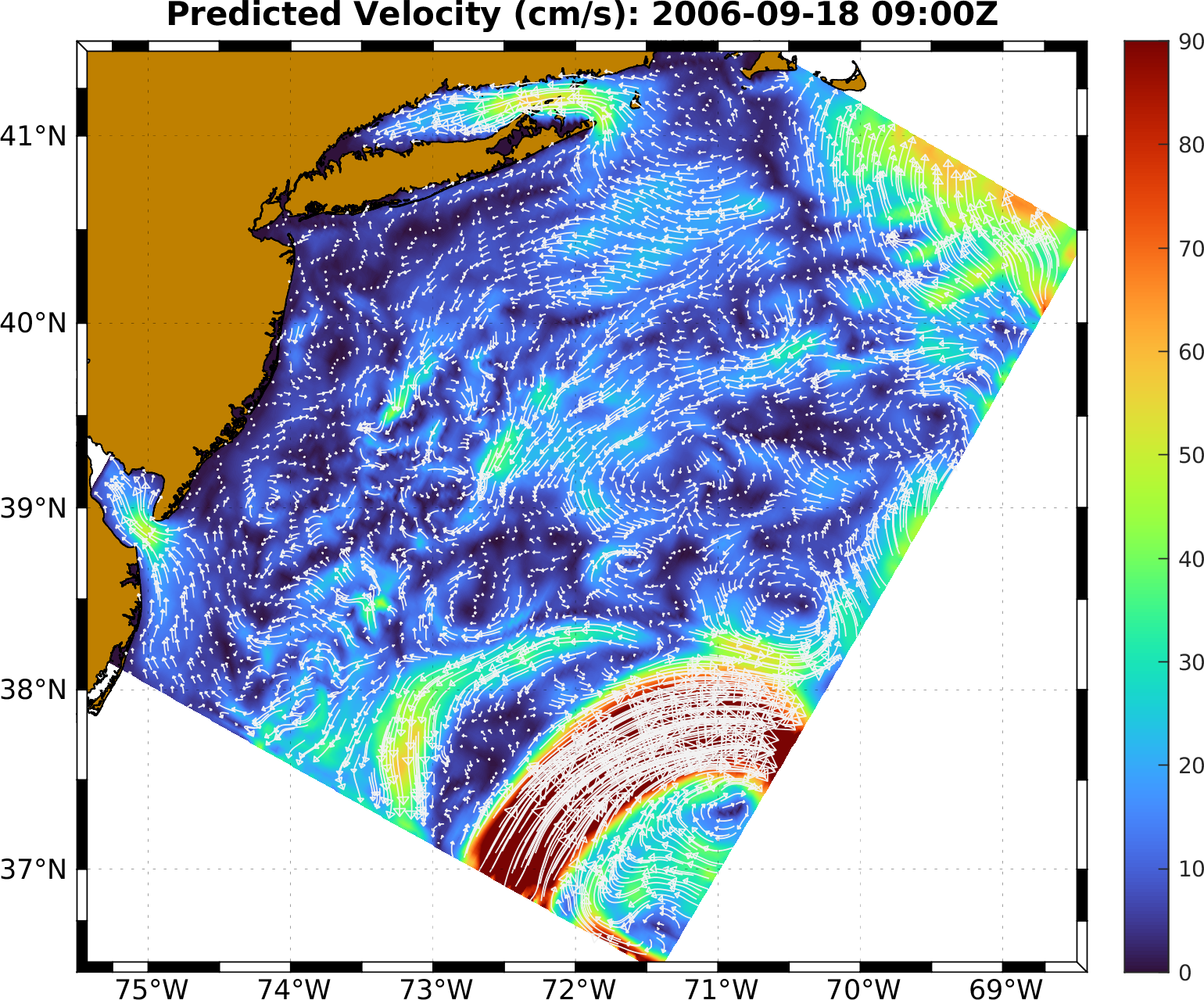}}
    \subfloat[]{\includegraphics[width=0.20\textwidth]{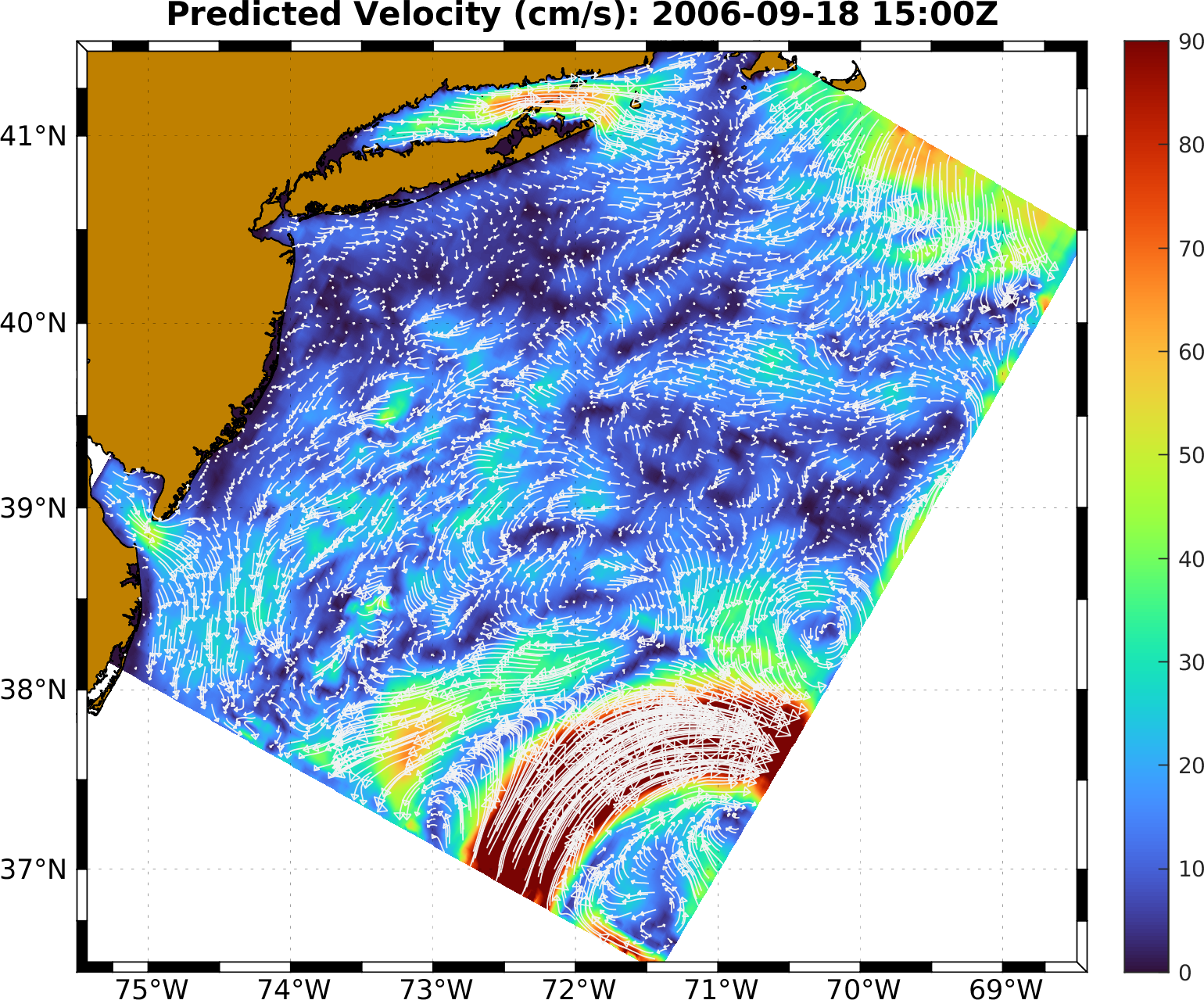}}
    \subfloat[]{\includegraphics[width=0.20\textwidth]{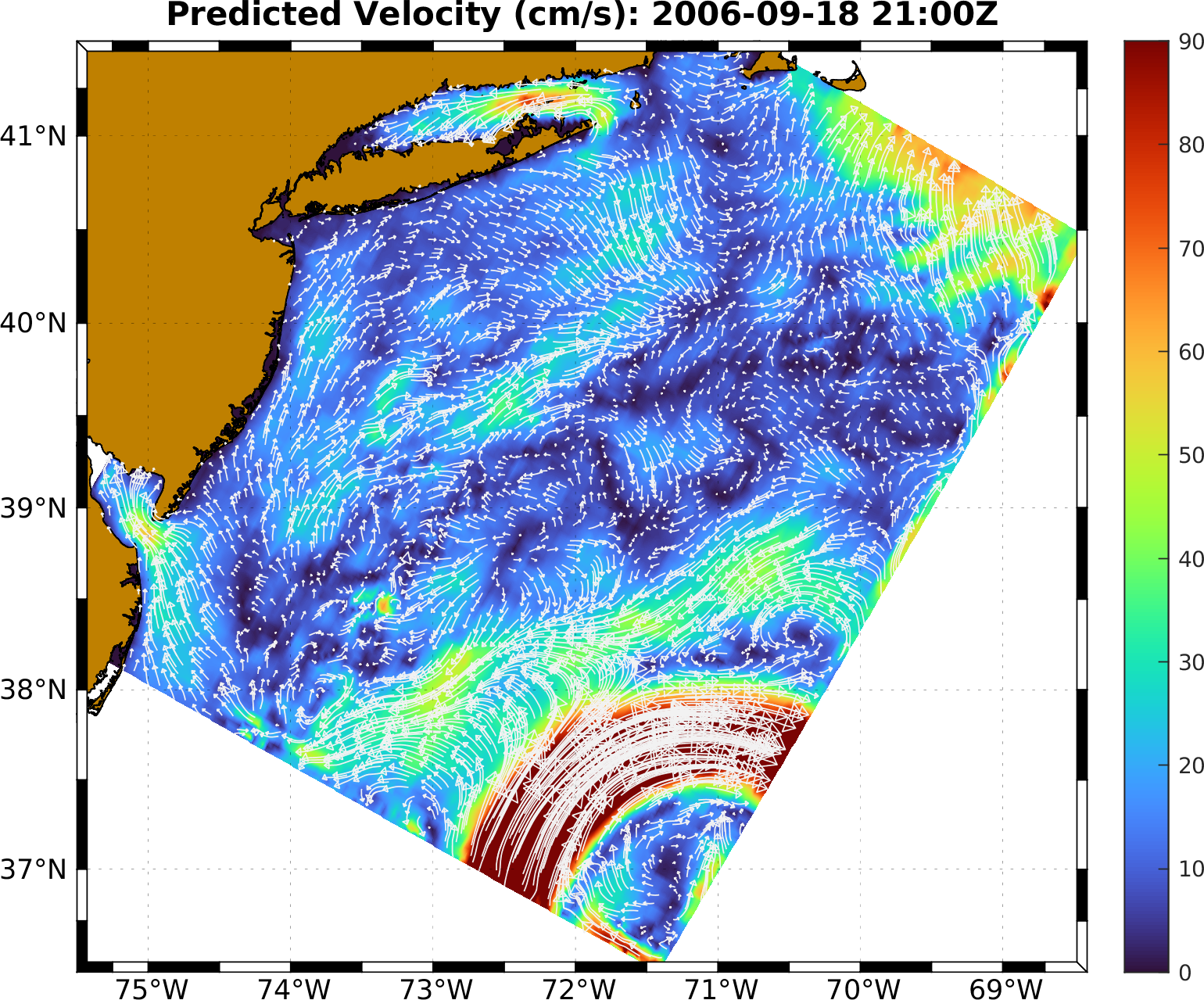}}
    \subfloat[]{\includegraphics[width=0.20\textwidth]{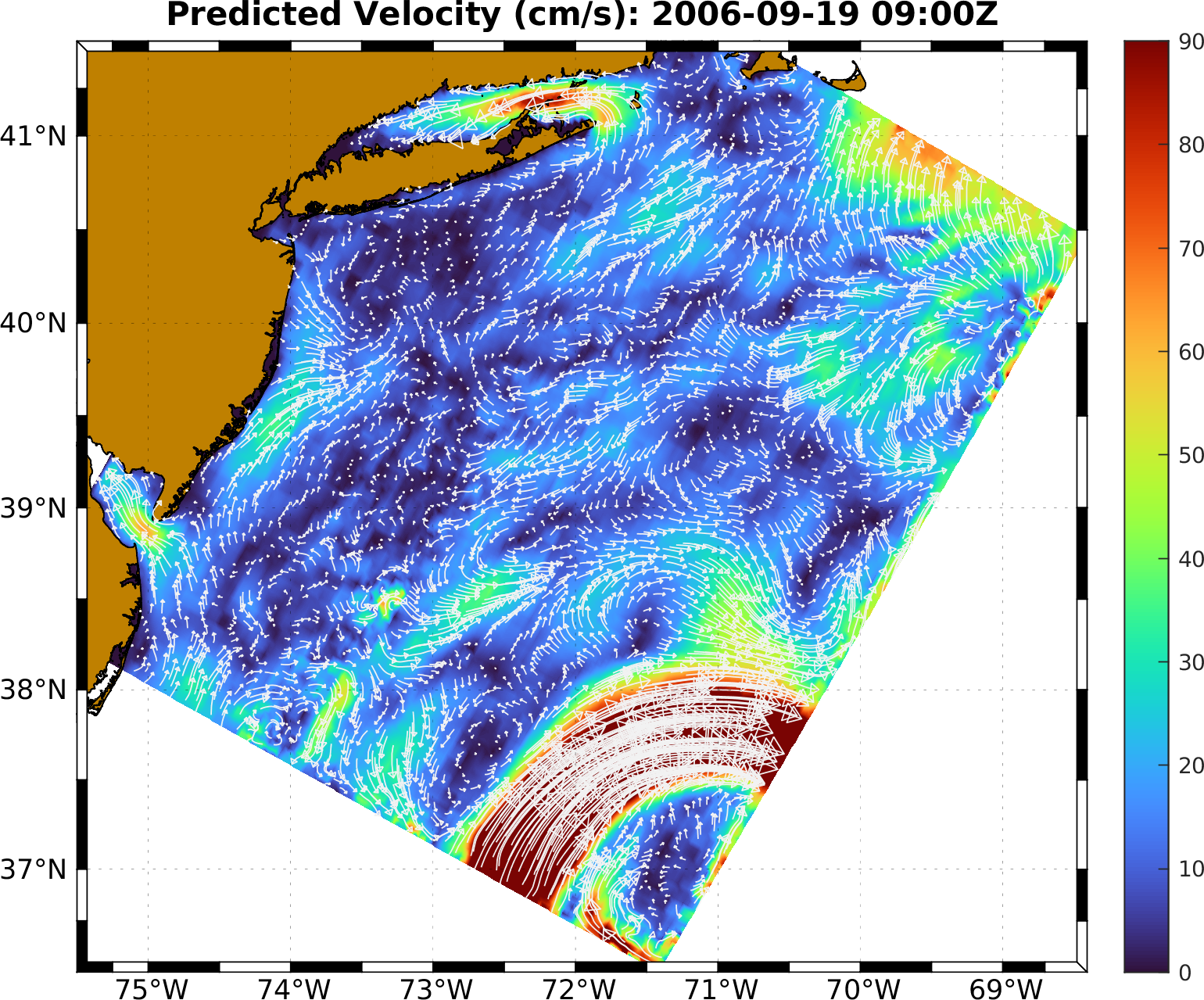}}
    \subfloat[]{\includegraphics[width=0.20\textwidth]{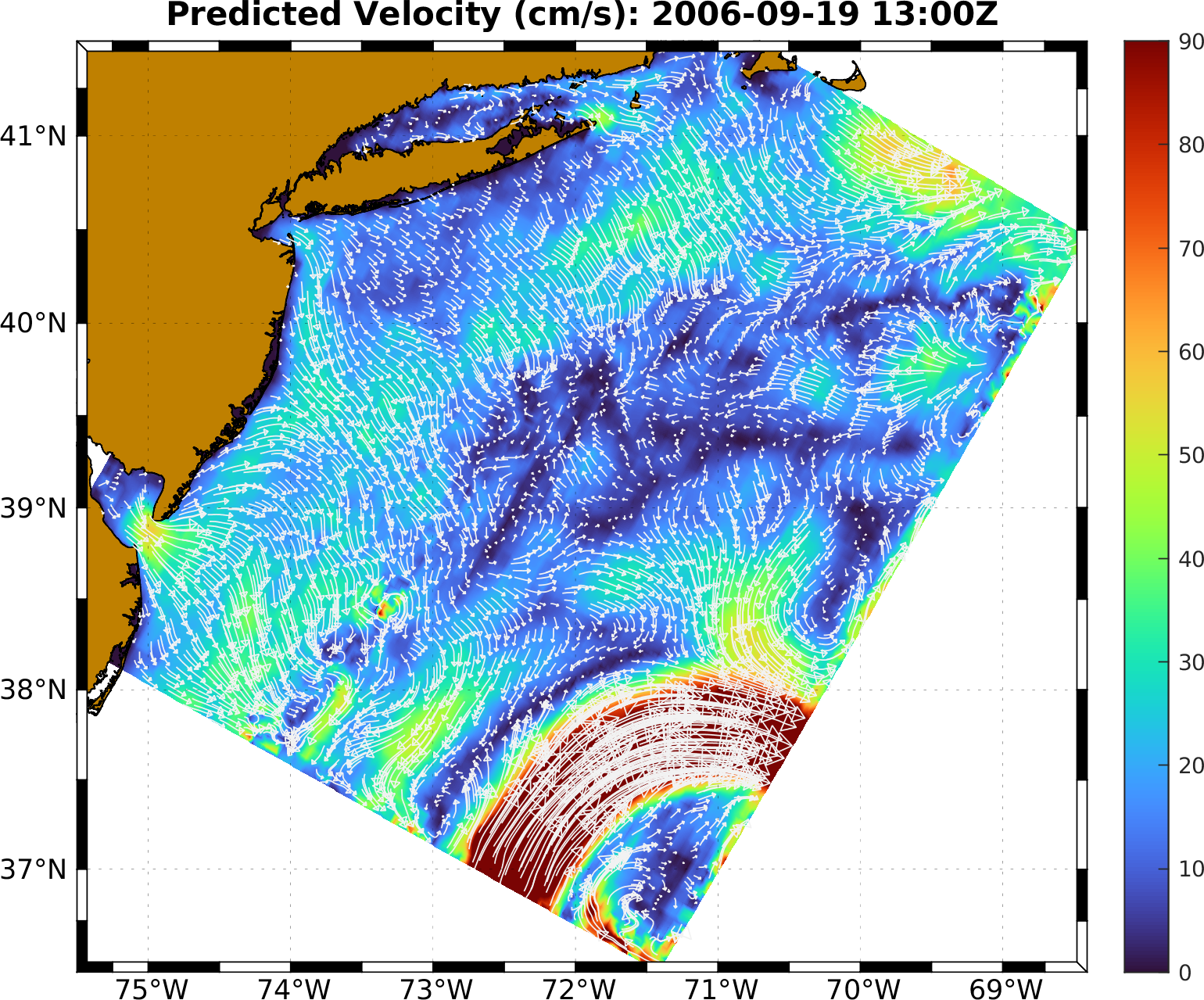}} \\
    \subfloat[]{\includegraphics[width=0.20\textwidth]{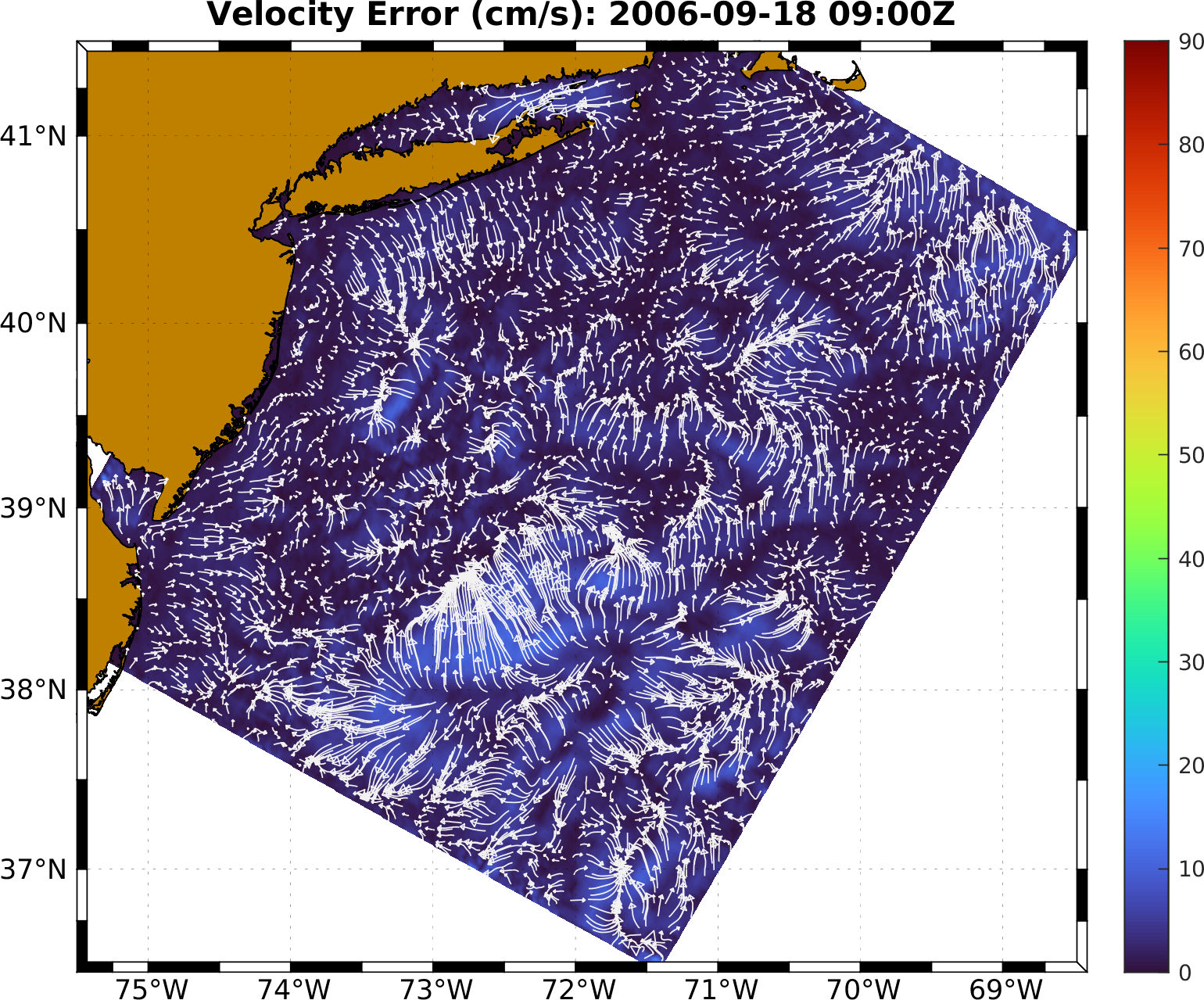}}
    \subfloat[]{\includegraphics[width=0.20\textwidth]{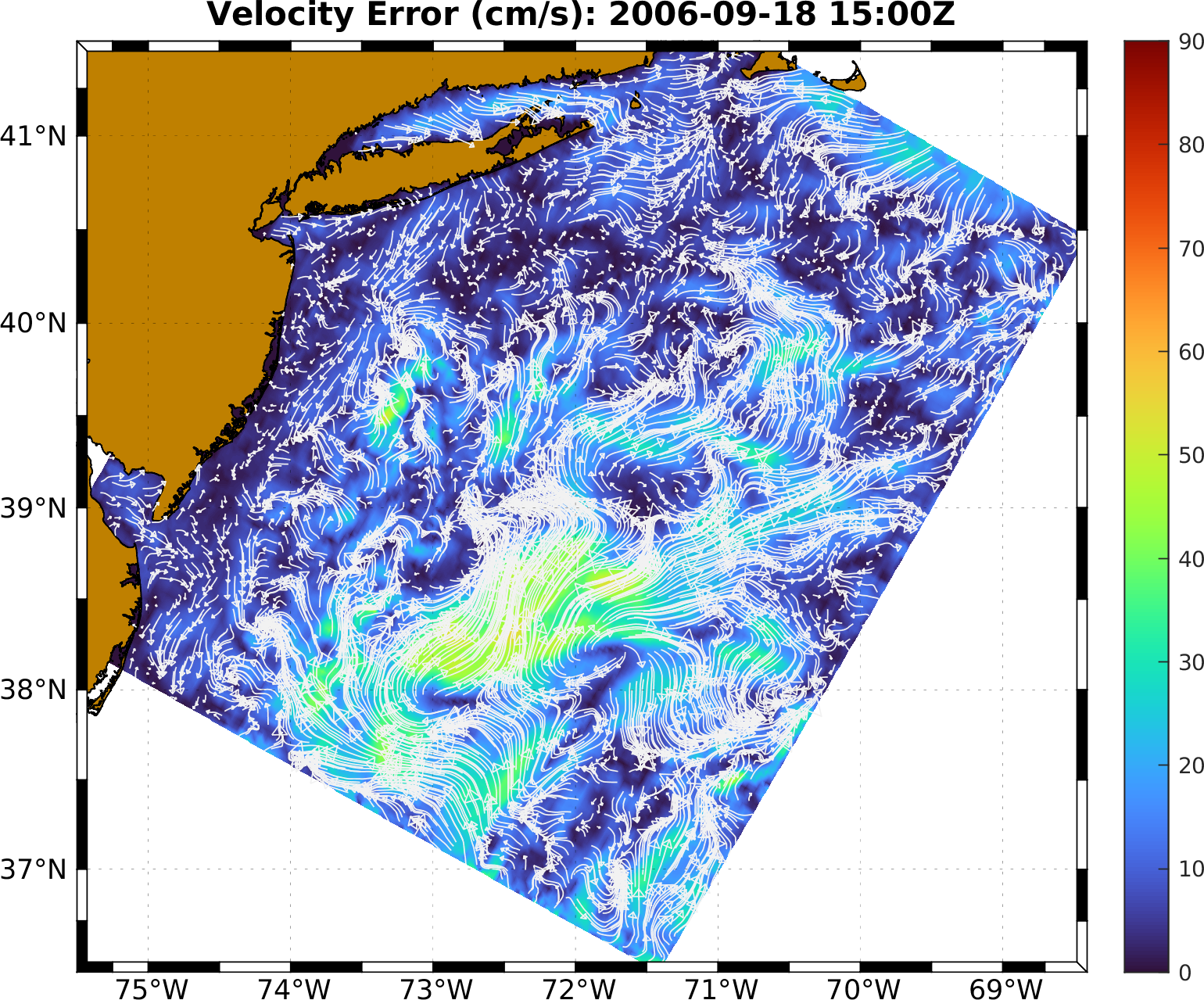}}
    \subfloat[]{\includegraphics[width=0.20\textwidth]{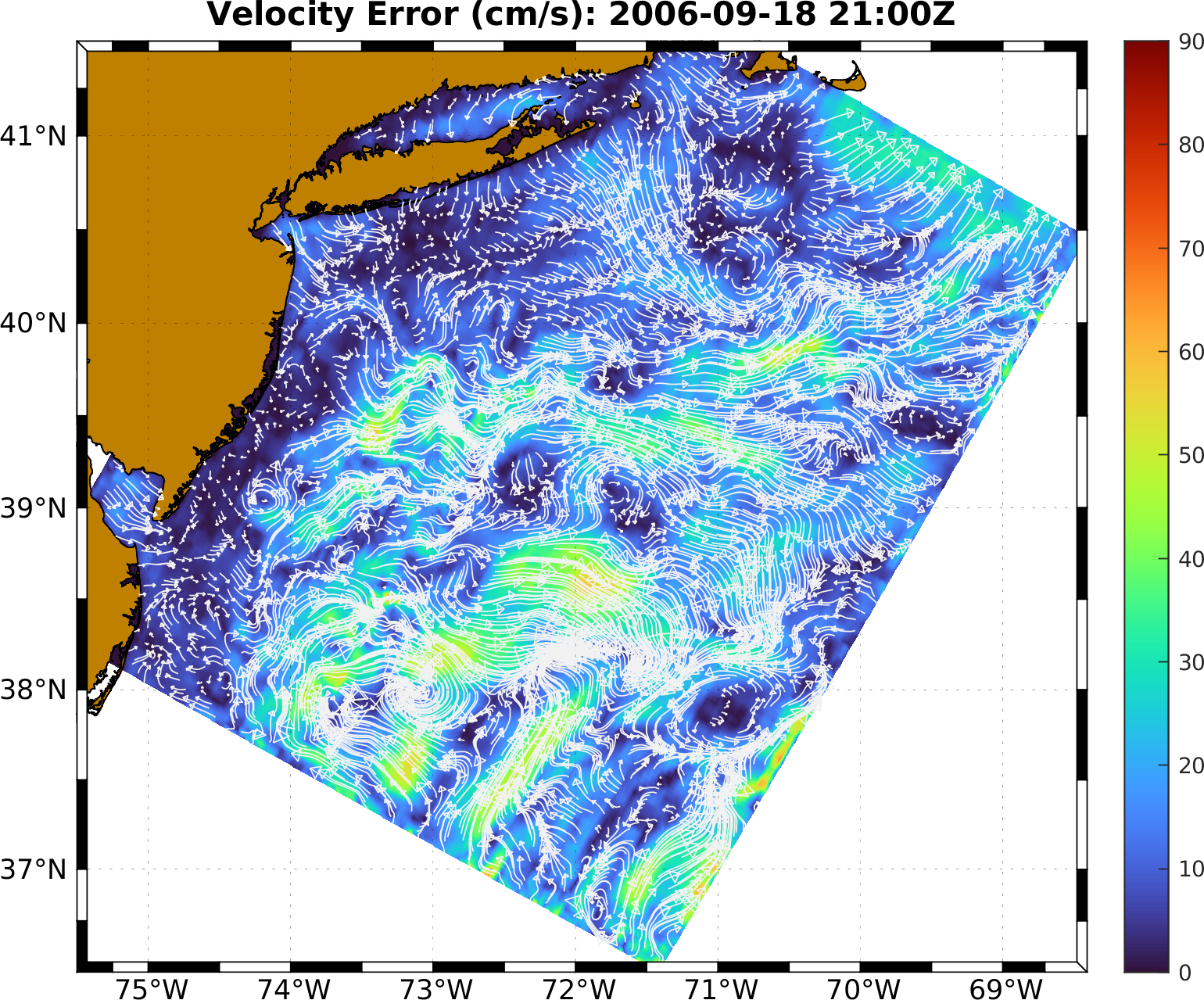}}
    \subfloat[]{\includegraphics[width=0.20\textwidth]{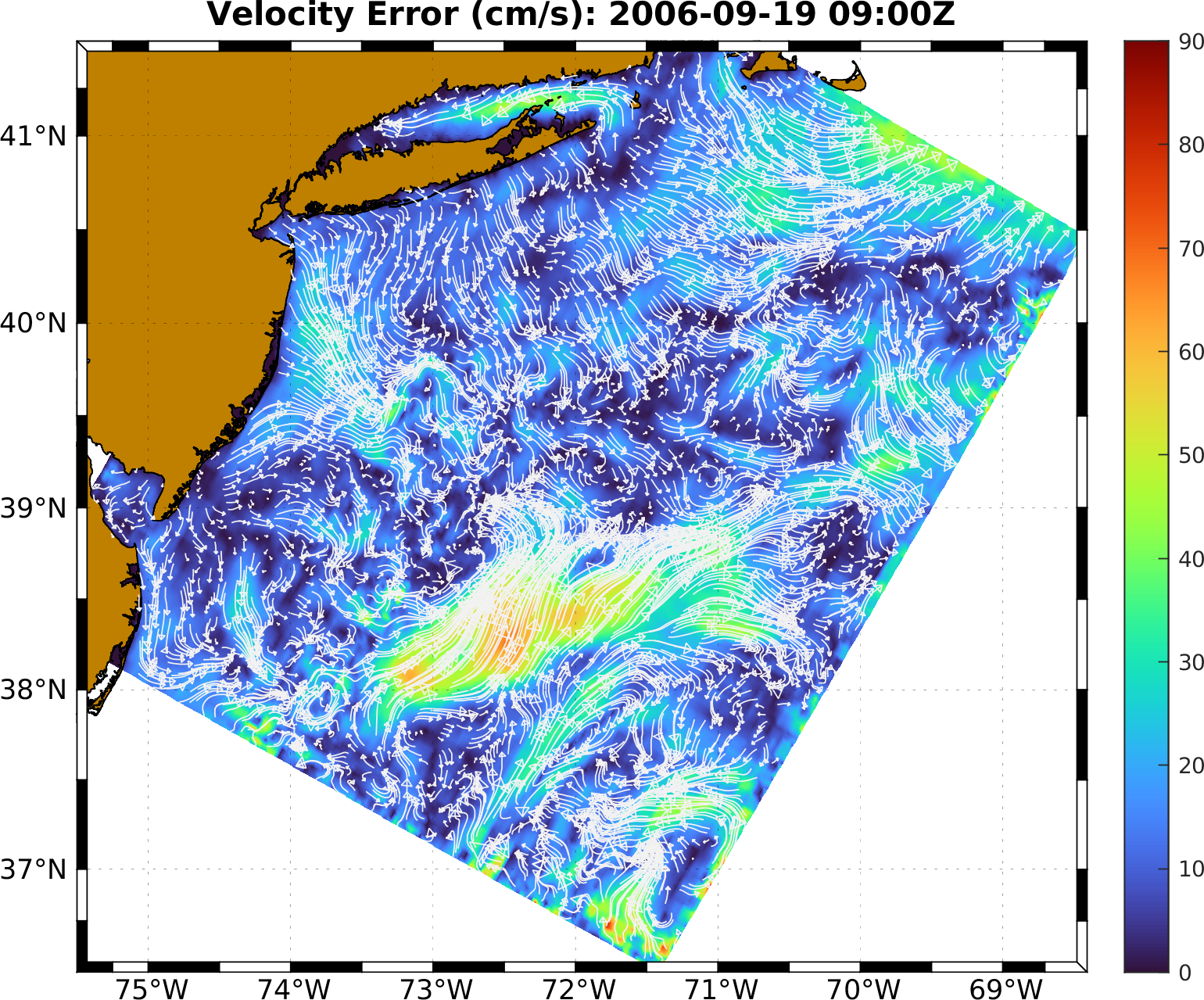}}
    \subfloat[]{\includegraphics[width=0.20\textwidth]{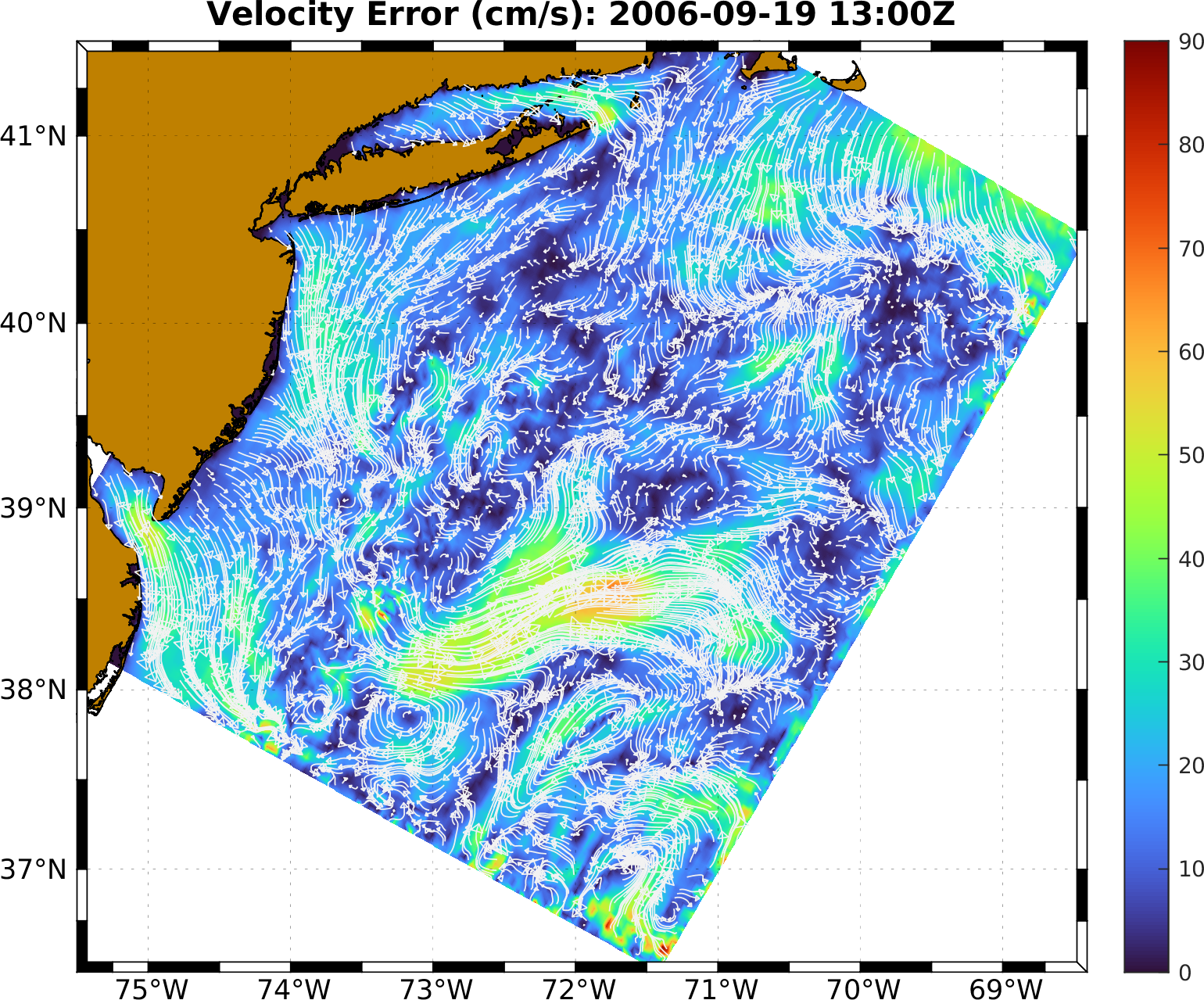}}
    \caption{\small \emph{Applying FCN in the MAB}. 
    Surface velocity field amplitudes overlaid with curved velocity vectors. Ground truth from the MSEAS-PE forecast (top row), FCN predictions for training-run 246 (middle row), and error fields (bottom row) for five different forecast times: 2006 Sep 18 09Z and 6, 12, 24, and 28 hours later.}\label{fig:awacs_fields}
\end{figure*}

\section{Realistic Ocean Surface Flows}

We now investigate the use of 
deep neural operator models to represent and forecast realistic surface ocean circulations in two ocean regions and dynamical regimes. We first examine shelf-to-deep dynamics in the Middle Atlantic Bight region with the Gulf Stream, shelfbreak front, internal and external tides, and large storms. We then consider shallow coastal dynamics in Massachusetts Bay with coastal currents and jets, strong tidal flows, and intermittent atmospheric forcing.

For these two applications, the learning data consist of high-resolution data-assimilative simulations computed for real sea experiments \cite{haley_et_al_Oceans2020,mseas_exercises_awacs_2006,haley_lermusiaux_OD2010} by our MIT Multidisciplinary Simulation, Estimation, and Assimilation Systems primitive equation (MSEAS-PE) modeling system. The MSEAS-PE has been used for fundamental research and for realistic simulations in varied regions of the World Ocean \cite{leslie_et_al_JMS2008,haley_et_al_DSR2009,lermusiaux_et_al_TheSea2017,subramani_et_al_JGR2017,kulkarni_et_al_Oceans2018,gupta_et_al_Oceans2019,lermusiaux_et_al_Oceans2019,ali_et_al_Oceans2023}.
It is able to simulate sub-mesoscale processes over regional domains with complex geometries and varied interactions using an implicit two-way nesting/tiling scheme \cite{haley_lermusiaux_OD2010,haley_et_al_OM2015}. Other capabilities include ensemble forecasting and data assimilation~\cite{lermusiaux_et_al_QJRMS2000,lermusiaux_JAOT2002,lermusiaux_DAO1999,lermusiaux_PhysD2007,lermusiaux_et_al_BBN_Oceans2020}, tidal prediction and inversion~\cite{logutov_lermusiaux_OM2008}, coastal objective analysis~\cite{agarwal_lermusiaux_OM2011}, subgrid-scale models~\cite{lermusiaux_JMS2001,lermusiaux_JCP2006}, 
reduced-order modeling
\cite{feppon_lermusiaux_SIREV2018,charous_lermusiaux_Oceans2021,heuss_et_al_Oceans2020,ryu_et_al_Oceans2021,ryu_et_al_Oceans2022},
and path planning and adaptive sampling \cite{lermusiaux_et_al_springer_HOE2016,lermusiaux_et_al_TheSea2017,lermusiaux_et_al_ST2007,lermusiaux_PhysD2007,doshi_et_al_CMAME2023}.

\subsection{Middle Atlantic Bight (MAB)}

We first evaluate the use of the FCN architecture based on a real ocean experiment in the MAB. The MAB is a broad shelf with strong tides, near-inertial wave activity, freshwater inputs, and a shelfbreak with a front.
It is influenced by the Gulf Stream, the warm-core rings, and a recirculating deep circulation, and is also affected by tropical storms \cite{tang_et_al_ocn2007,gangopadhyay_et_al_JAOT1997}.

\subsubsection{Test Case Description}
\label{sec:test_case_MAB}

%
%

%
The MAB simulation data are an outgrowth of the real-time MSEAS-PE forecasts issued during the Shallow Water 2006 (SW06) and Autonomous Wide Aperture Cluster for Surveillance (AWACS) 2006 experiments \cite{tang_et_al_ocn2007,chapman_lynch_JOE2010,mseas_exercises_awacs_2006,haley_lermusiaux_OD2010,lin_et_al_JOE2010,colin_et_al_OCEANS2013}.
The MSEAS-PE was configured in a two-way nested mode with a 3\;km resolution regional domain ($150\times175$ horizontal grid) and a 1\;km resolution shelfbreak domain ($173\times156$ horizontal grid). Both domains employ 100 terrain-following vertical levels optimized for the background T/S structure of the slope/shelf region. Surface forcing was a combination of the 6\;km Weather Research \& Forecasting (WRF) \cite{ruwrf_2020} and the $1^\circ$ Navy Operational Global Atmospheric Prediction System (NOGAPS) \cite{nogaps_2020} atmospheric fluxes. 
The nested simulations were forced with tides from the high resolution TPXO8-Atlas \cite{EgbertErofeeva2002,EgbertErofeeva2013} adjusted for our higher resolution bathymetry and coastlines \cite{logutov_lermusiaux_OM2008}.
The final data-assimilative reanalysis spanned Aug.\;14 to Sep.\;24, 2006, and was saved at hourly resolution providing a total of~$1007$ snapshots.
Additional details can be found in \cite{haley_lermusiaux_OD2010,kelly_lermusiaux_JGR2016,pan_et_al_JFM2021}.

The FCN results presented in this paper are for the surface circulation in the regional (3\;km) domain.
The first 48 snapshots are not used (dynamical adjustments) and 959 snapshots of size 2x150x174 are available (each snapshot is an array for $u$ and $v$ in cm/s). The snapshots contain both land and ocean regions with 4,433 land cells and 21,667 ocean cells.
%

\emph{ML data usage, NN architecture, loss function, parameters, and sensitivity studies}.
%
For what we show, we 
employ the first $600$ snapshots of surface velocity $u$ and $v$ for training, the next $200$ for validation, and the remaining 159 for verification/prediction.
We completed sensitivity studies for these periods, as well as for several of the FCN architectural choices and key hyperparameters such as the loss function, number of epochs, global batch size, size of image patches (p$\times$p), number of AFNO layers (depth), and embedded channel size.
For example, we varied the global batch size from 1 to 12, the image patch size from 2x2 to 6x6, the depth from 10 to 25, and the embedded channel size from 384 to 768. 
The model was trained for about 150 to 200 epochs.
For the field results shown next, the 
MAB-trained FCN model corresponds to MAB-training-run\;246 that has a tuned global batch size of 1, size of image patches of 3x3, number of AFNO layers (depth) of 10, and embedded channel size of 640. 
For this training-run\;246, the other hyperparameters such as learning rate, sparsity threshold, and MLP ratio were those of the original FCN model \cite{pathak2022fourcastnet}. 
The loss function employed was the L2-norm.
To capture the dynamics, we found that tuning the patch size and increasing the embedded channel size led to the best results during training, validation, and prediction.

\subsubsection{Learning Results}

%



A comparison of the simulated ground truth, an MSEAS-PE forecast of the surface velocity field for 2006 Sep 18-19 in the MAB, to an FCN prediction 
trained on earlier MSEAS-PE fields (training-run 246) is shown in Figure\;\ref{fig:awacs_fields}.  
The Gulf Stream is present
in the southern corner of the domain, with a small crest that begins to move out of the domain. This is
a period of low winds so the shelf is dominated by tides (primarily M2) 
 especially on Nantucket Shoals
(northern boundary) and internal tides.
The FCN forecast captures the changes in the Gulf Stream well. It also captures predominantly
M2 tidal signal on the shelf, 
although it slightly overestimates the tidal strength.

\begin{figure}[h!]
    {\includegraphics[width=\linewidth]{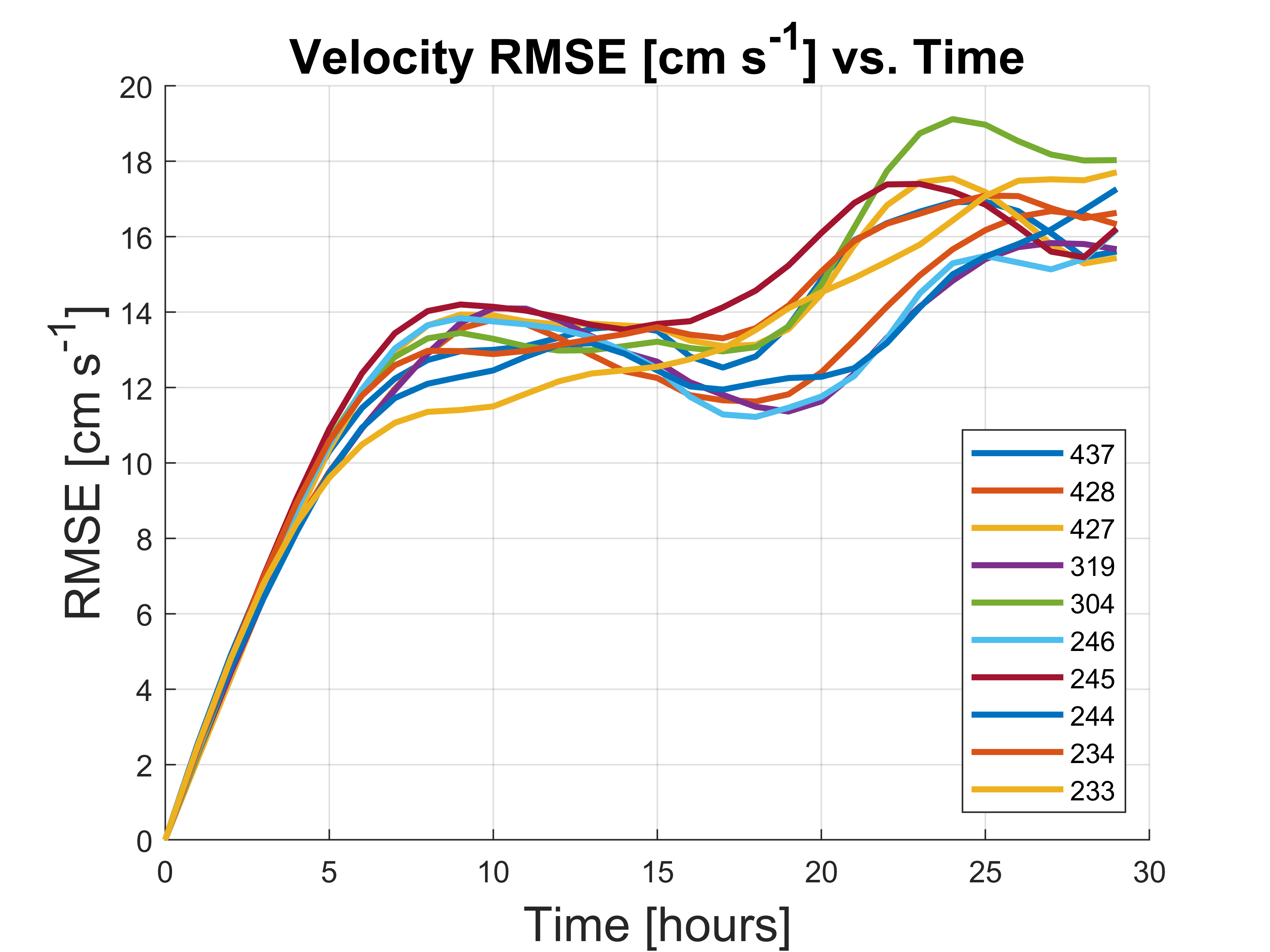}}
        \caption{\small \emph{Applying FCN in the MAB}. Errors (RMSE) of FCN forecasts of surface velocity for 0 to 29h forecast lead times (2006 Sep 18 08Z to Sep 19 13Z). The FNC forecasts correspond to ten MAB training-runs with different hyperparameter values.
        }
    \label{fig:awacs_1D_overlay}
\end{figure}

In Figure\;\ref{fig:awacs_1D_overlay}, we show the root-mean-square-errors (RMSEs) of ten FCN forecasts of MAB surface velocity fields for 0 to 29h forecast lead times. The different FCN forecasts were selected among the better ones (one of which is MAB-training-run 246 shown on Fig.\;\ref{fig:awacs_fields}) and correspond to ten MAB training-runs with different hyperparameter values.
For these FCN forecasts,
averaged errors grow linearly quickly, and then taper off or grow more slowly, with 
a clear effect of M2 tides.
Results confirm that the FCN forecasts have accuracy, with local errors reaching about 5 to 50\% of the variability after 29\;h.  The FCN forecast uncertainty is estimated by the spread of the RMSE curves shown.  

\subsection{Massachusetts Bay (MB)}

\begin{figure*}
    \centering
    \subfloat[]{\includegraphics[width=0.20\textwidth]{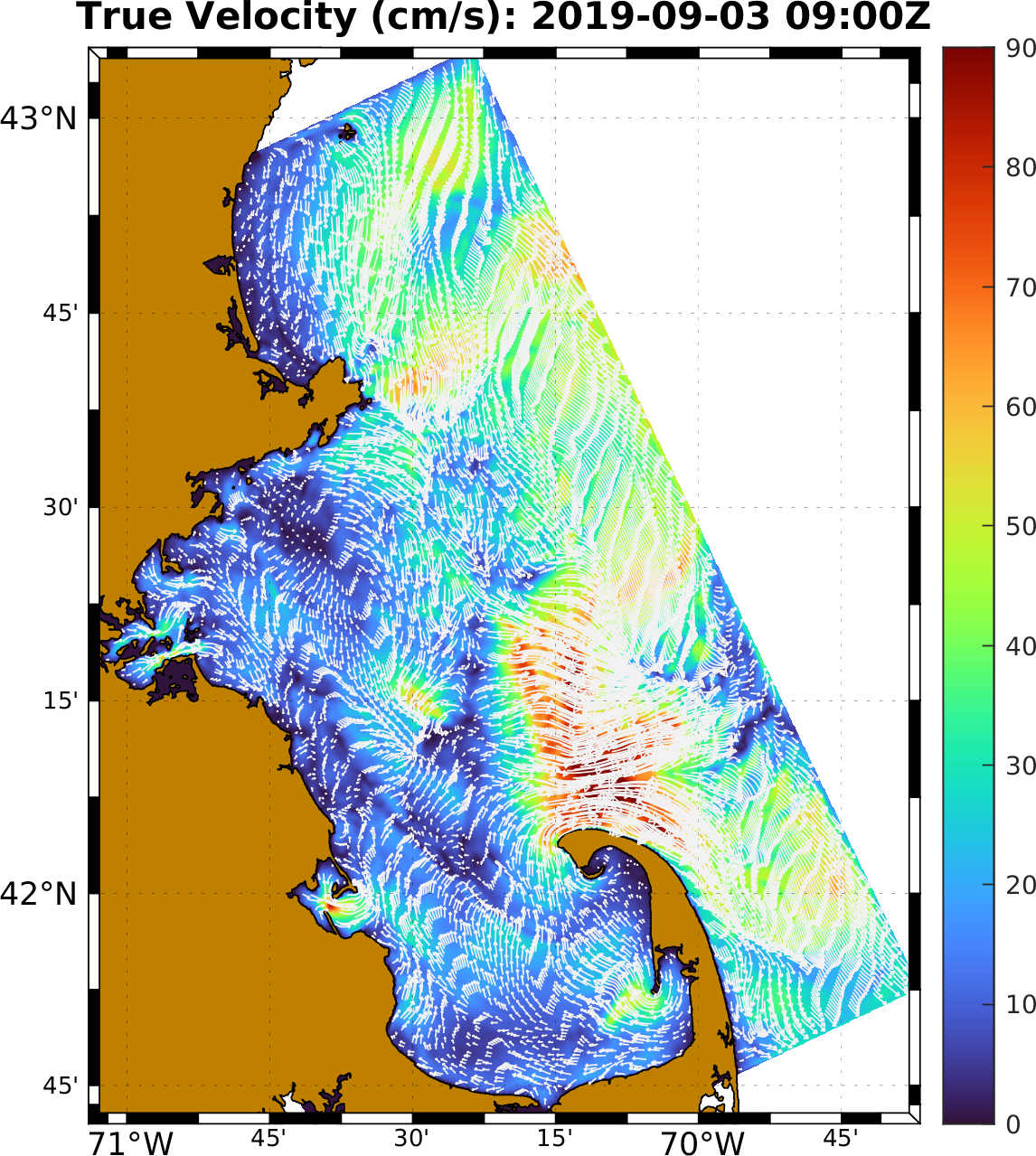}}
    \subfloat[]{\includegraphics[width=0.20\textwidth]{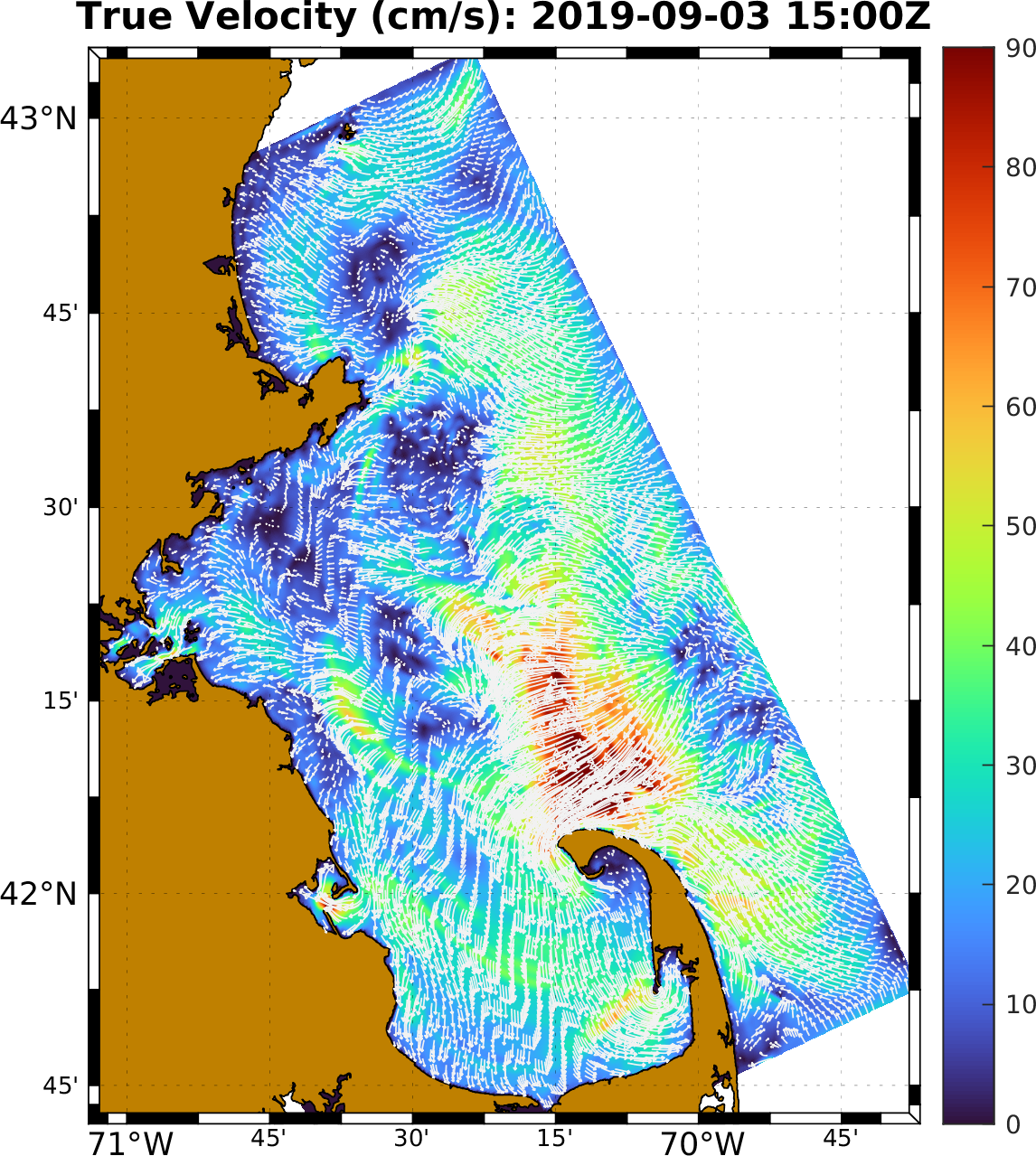}}
    \subfloat[]{\includegraphics[width=0.20\textwidth]{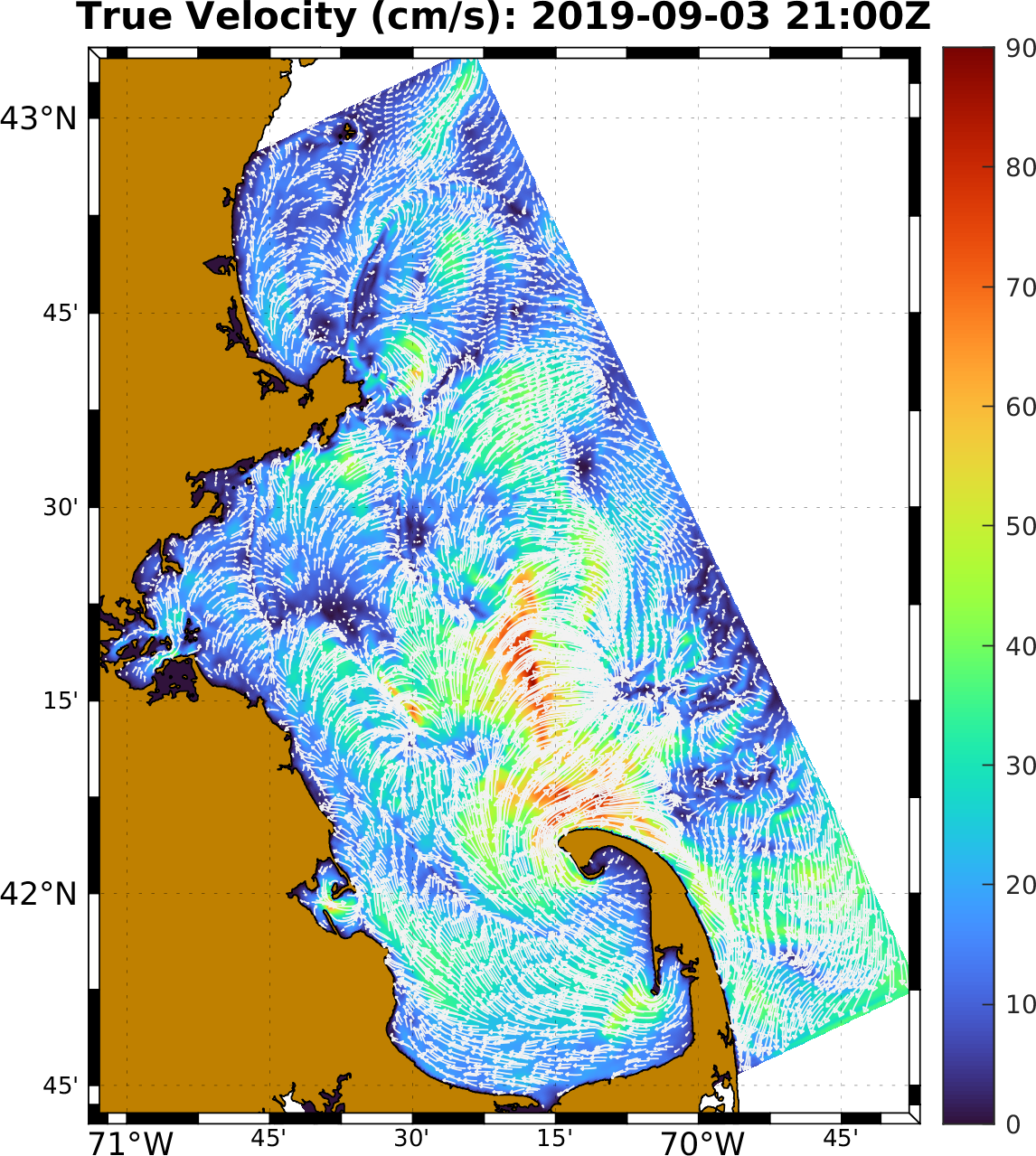}}
    \subfloat[]{\includegraphics[width=0.20\textwidth]{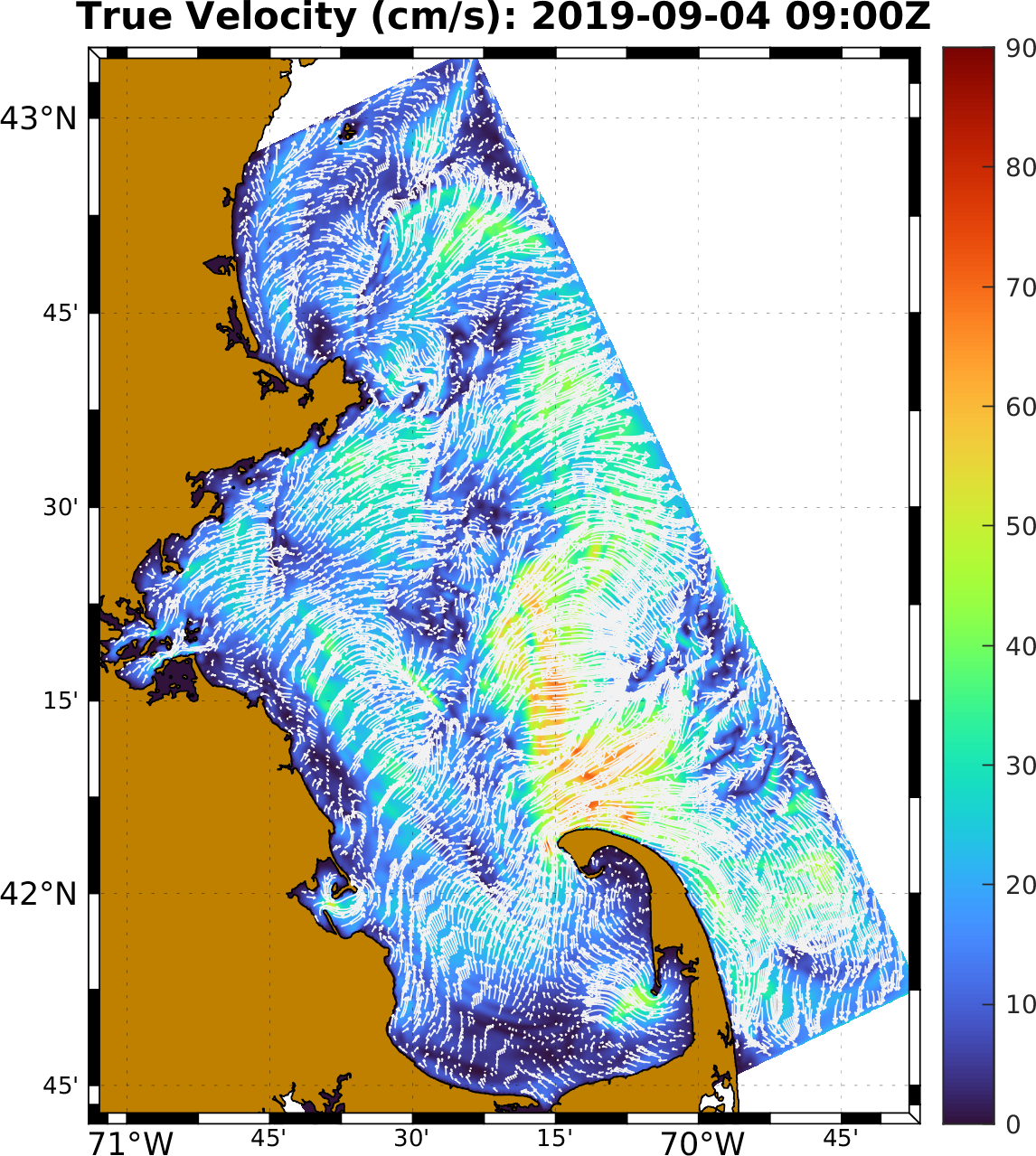}}
    \subfloat[]{\includegraphics[width=0.20\textwidth]{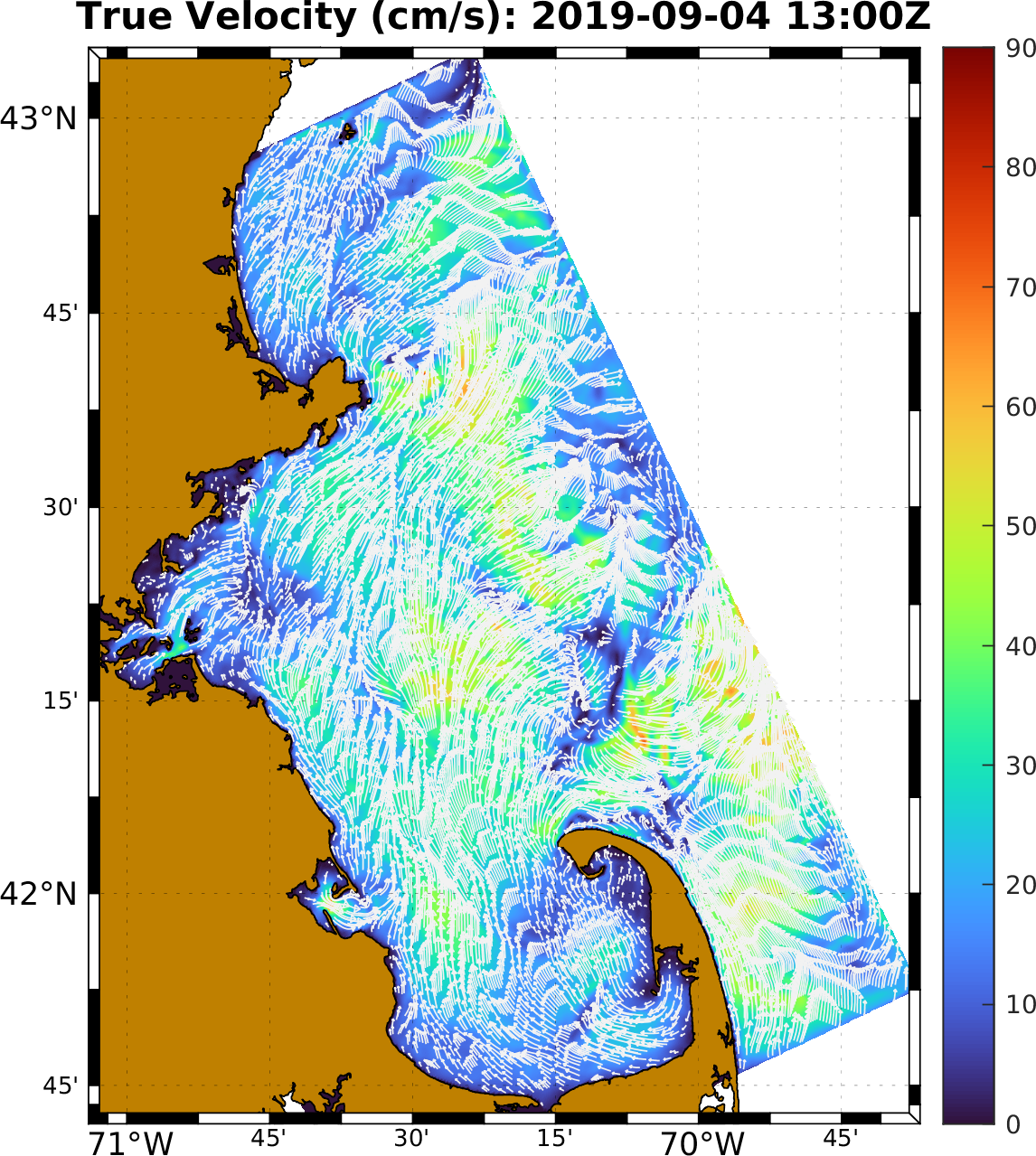}} \\
    \subfloat[]{\includegraphics[width=0.20\textwidth]{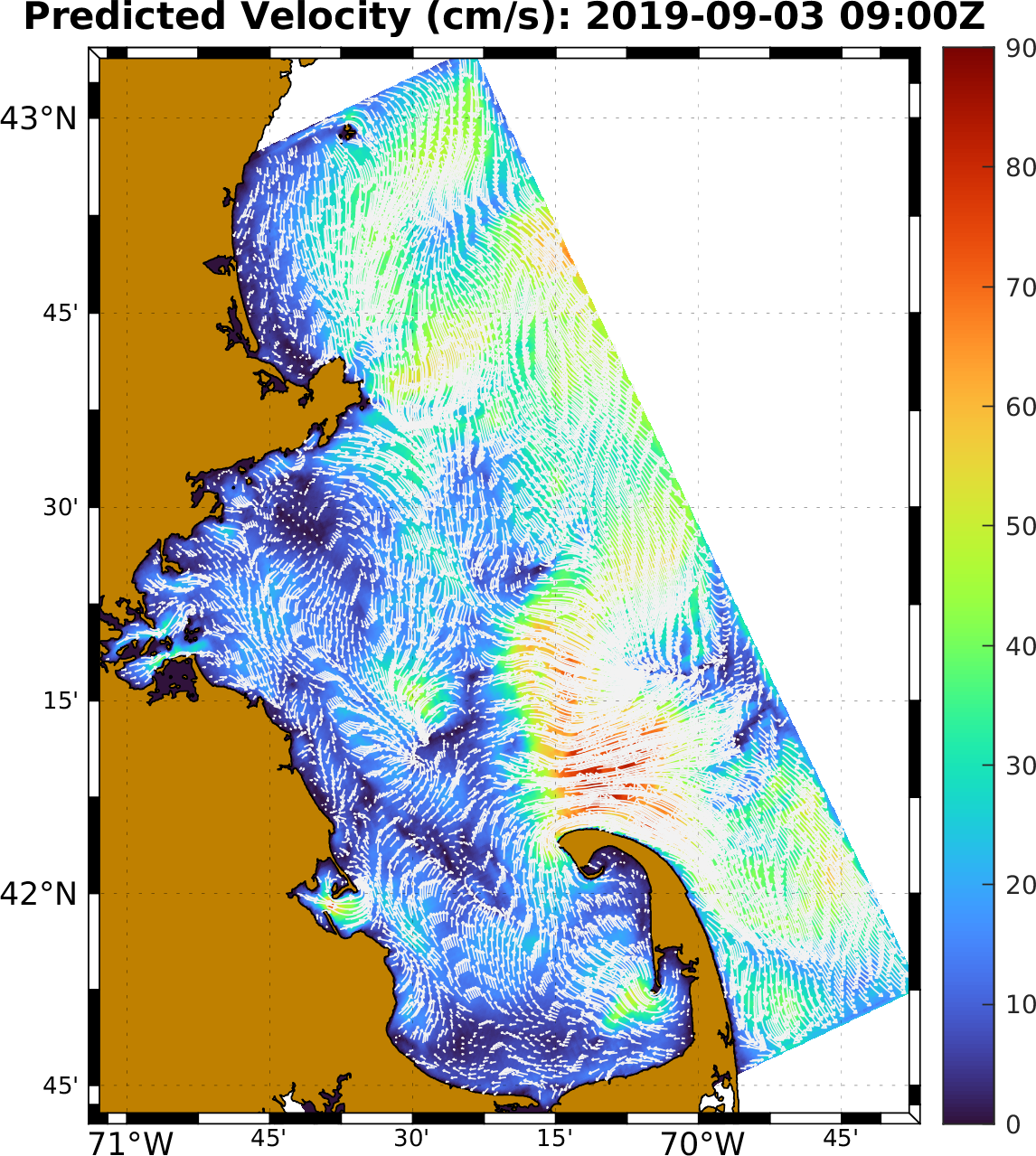}}
    \subfloat[]{\includegraphics[width=0.20\textwidth]{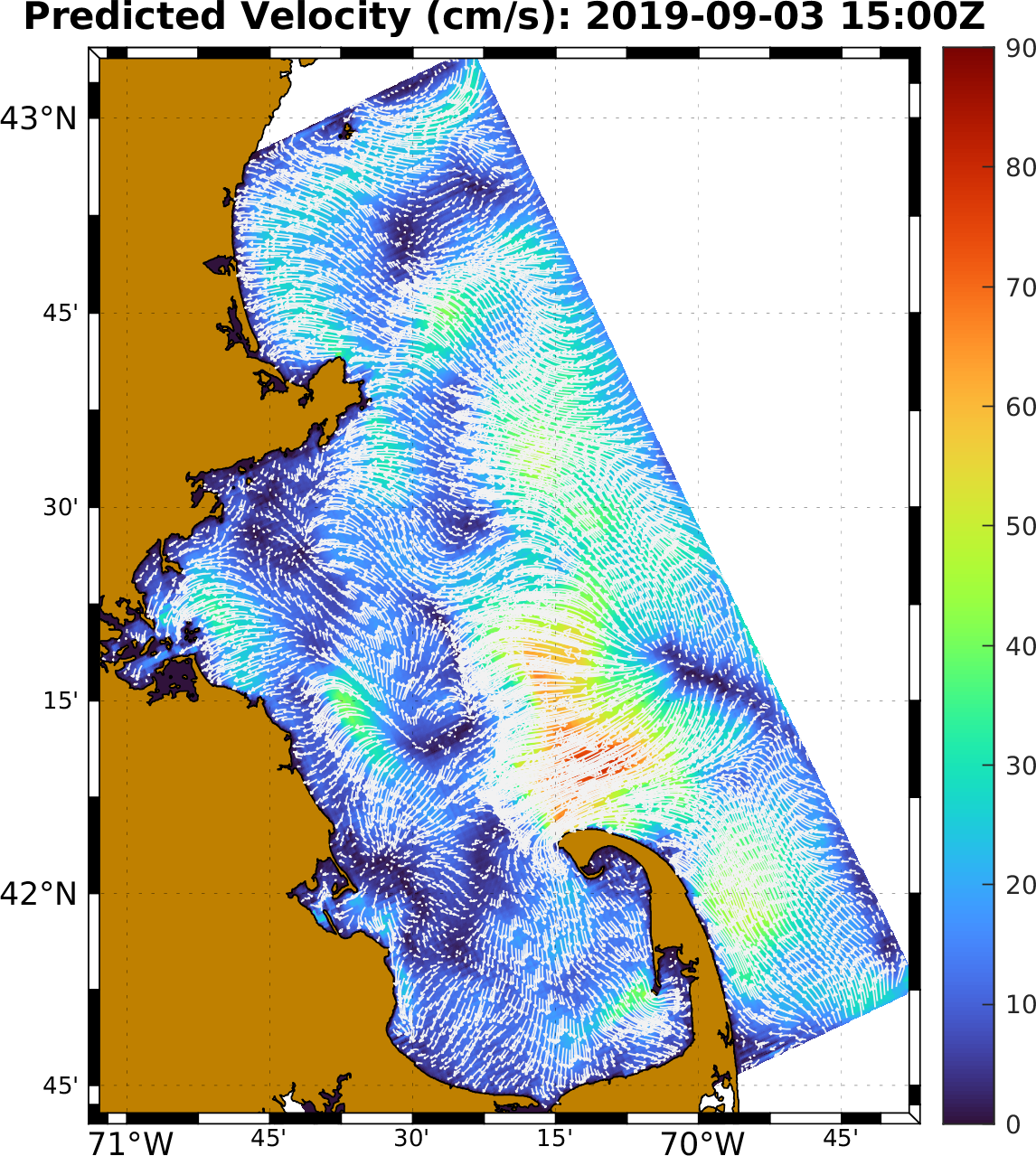}}
    \subfloat[]{\includegraphics[width=0.20\textwidth]{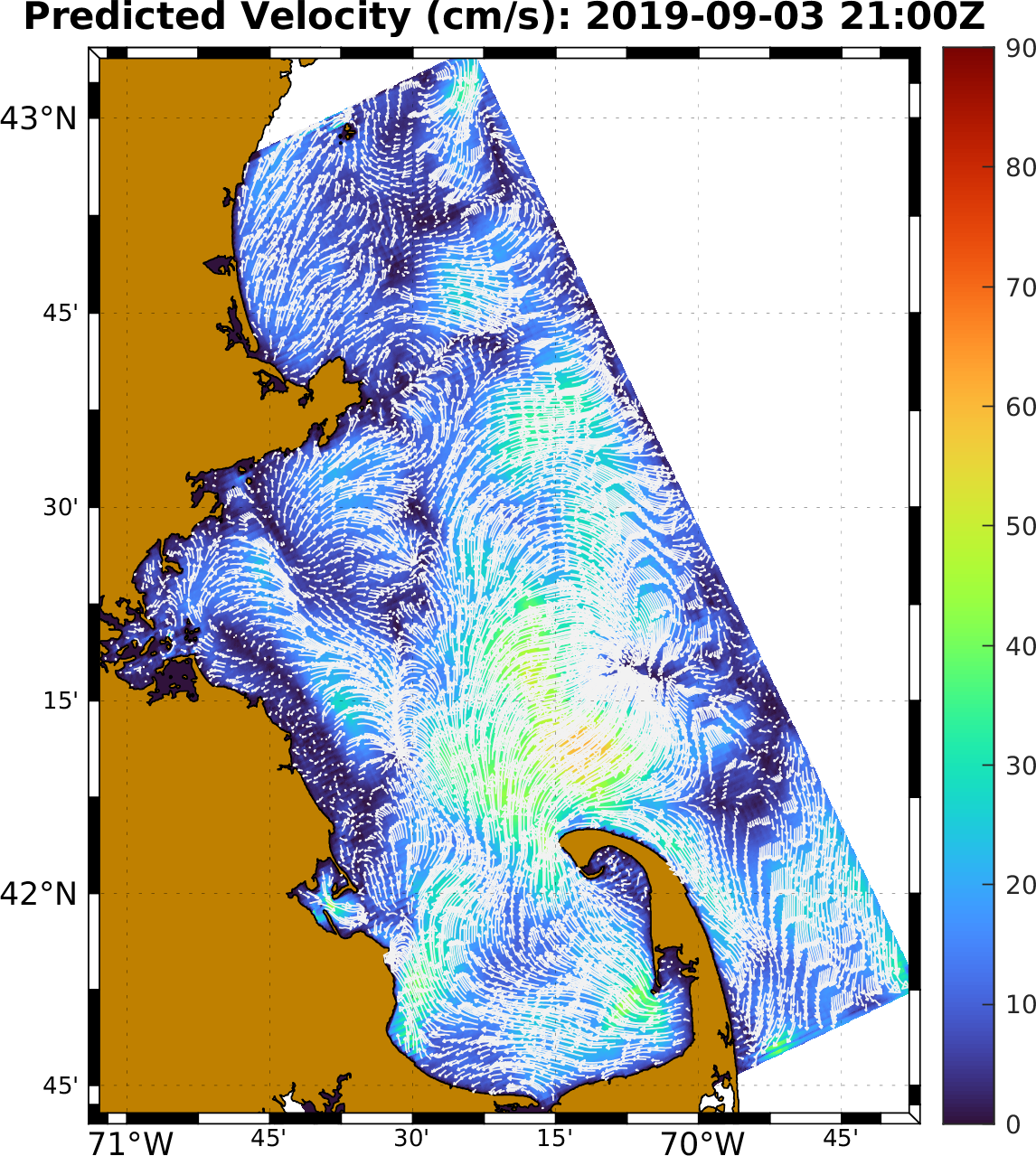}}
    \subfloat[]{\includegraphics[width=0.20\textwidth]{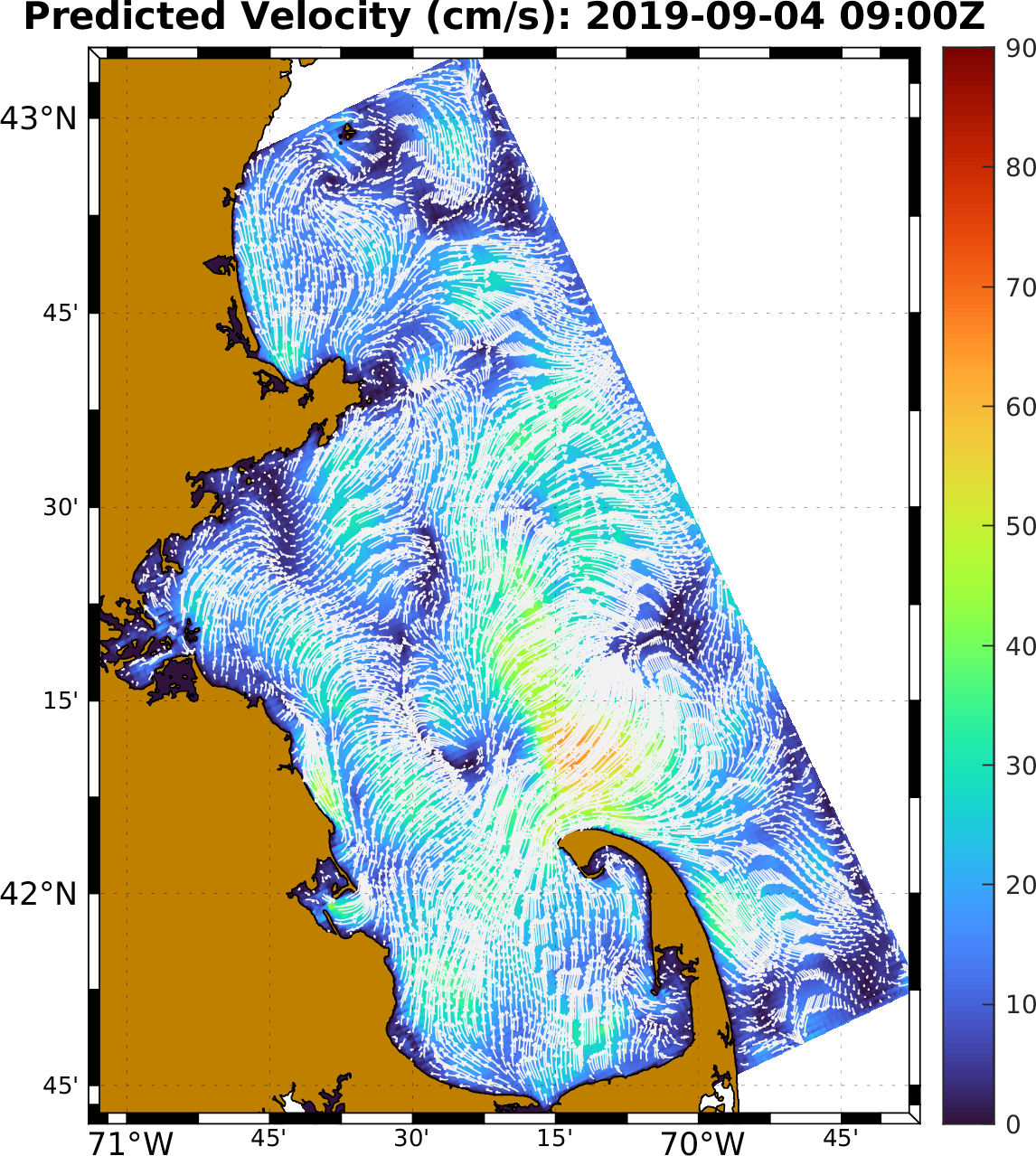}}
    \subfloat[]{\includegraphics[width=0.20\textwidth]{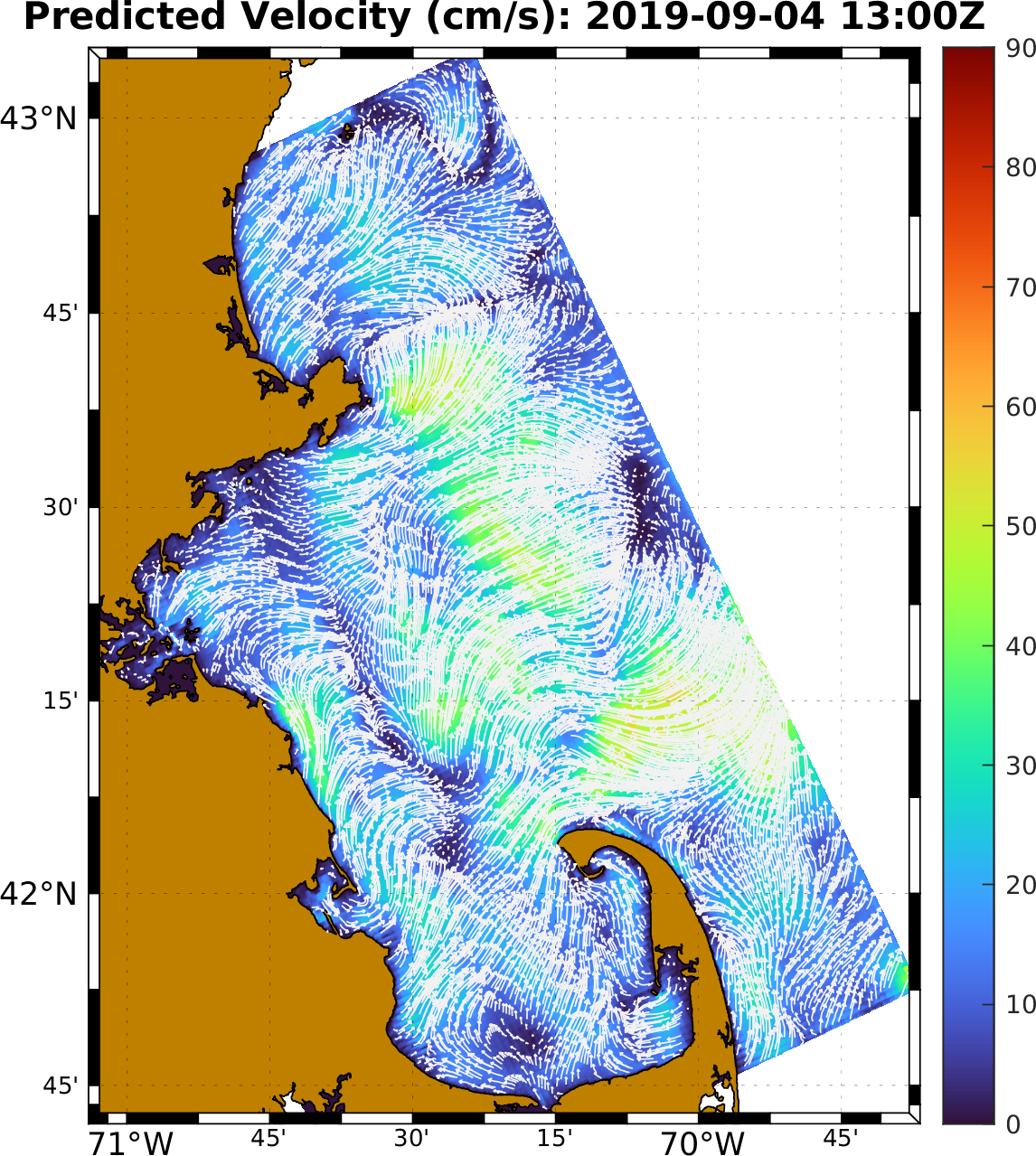}} \\
    \subfloat[]{\includegraphics[width=0.20\textwidth]{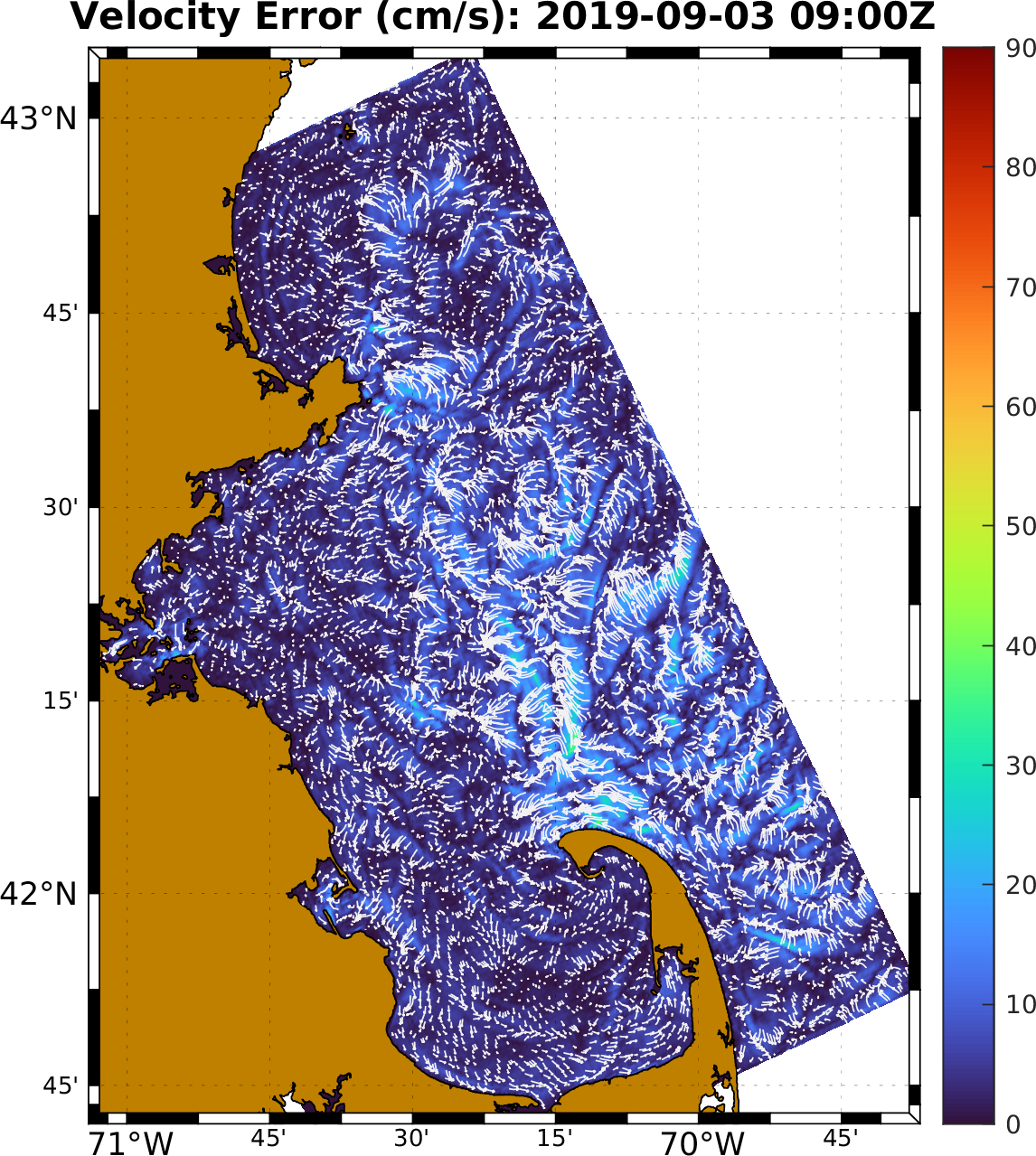}}
    \subfloat[]{\includegraphics[width=0.20\textwidth]{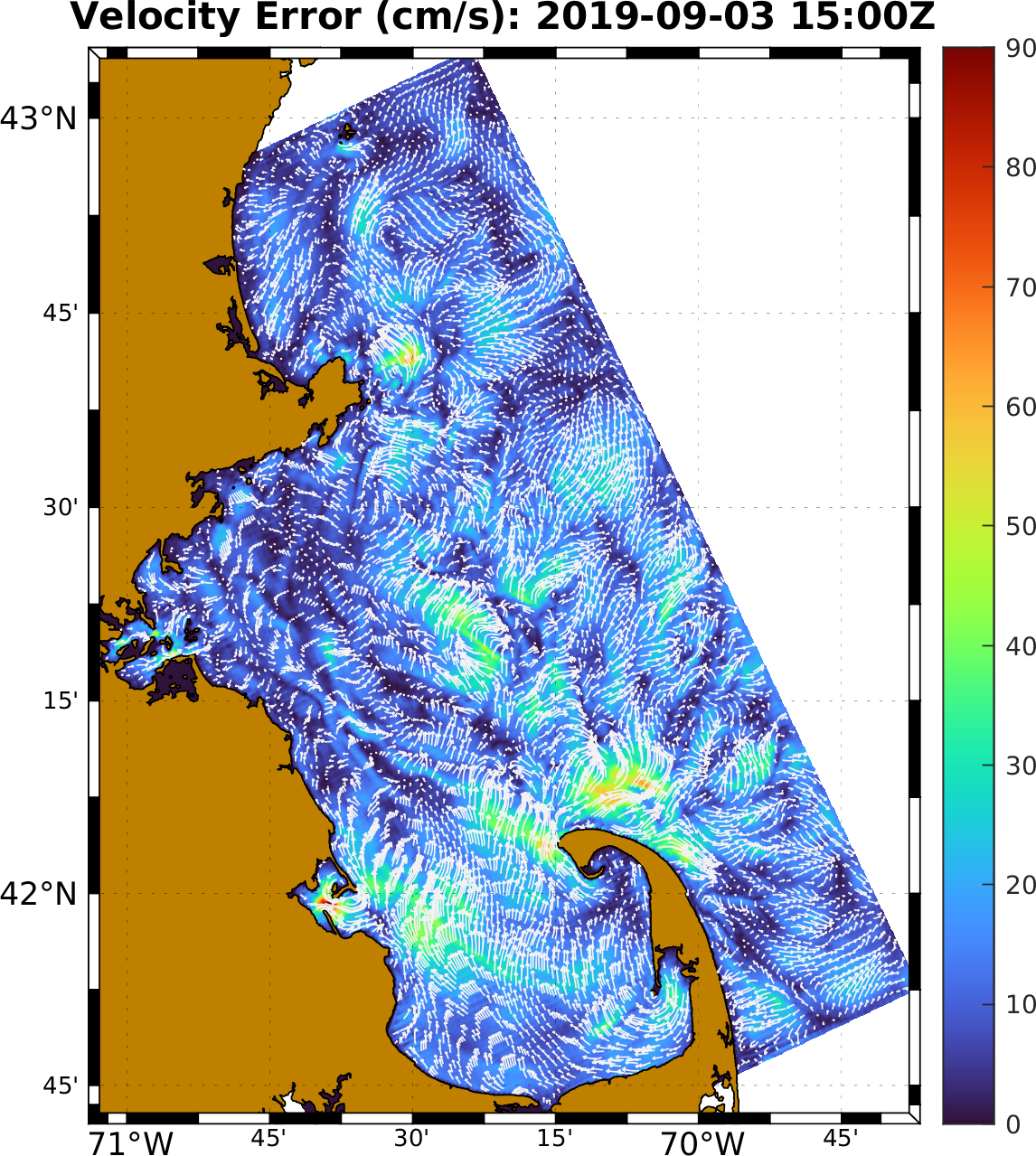}}
    \subfloat[]{\includegraphics[width=0.20\textwidth]{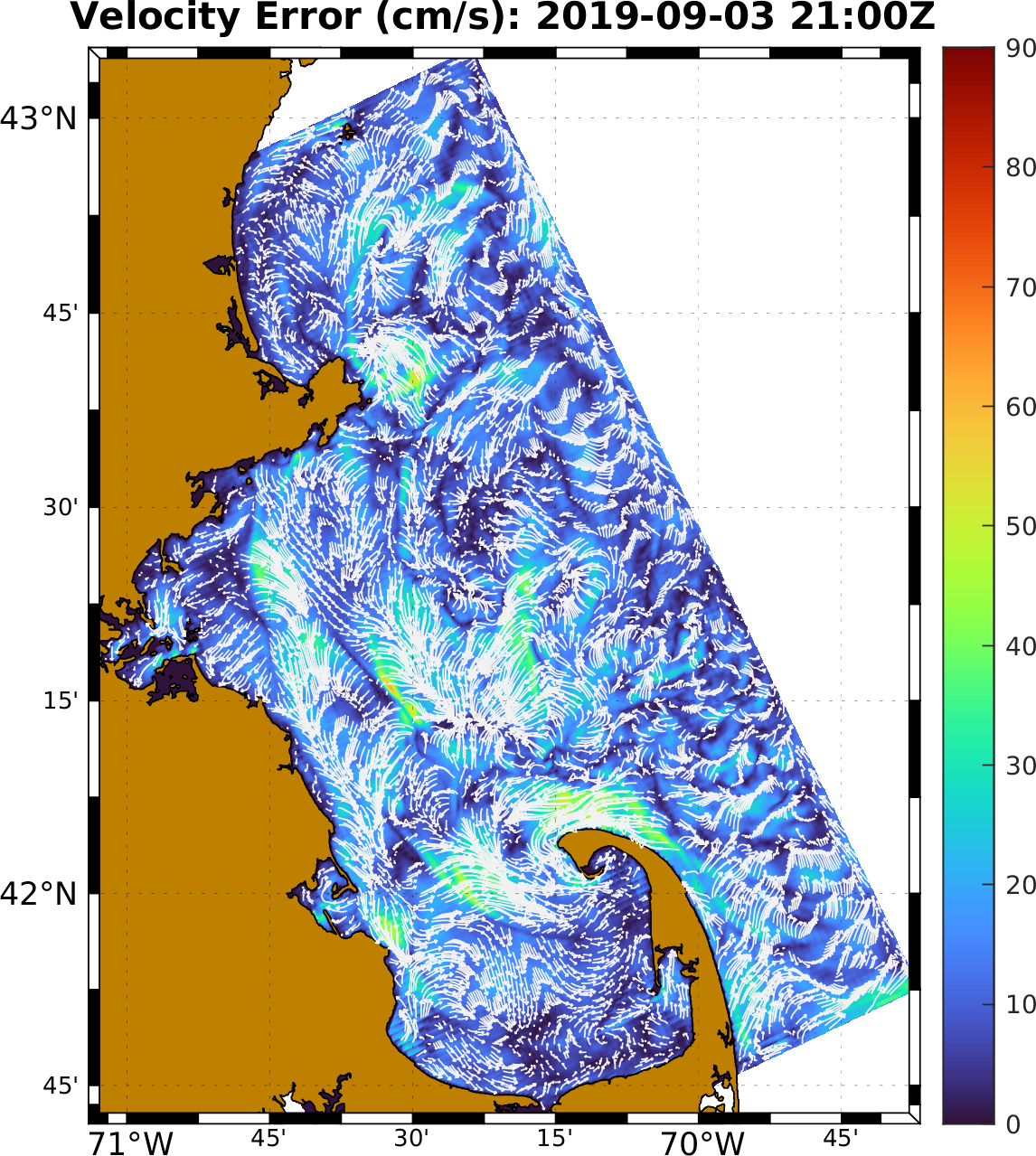}}
    \subfloat[]{\includegraphics[width=0.20\textwidth]{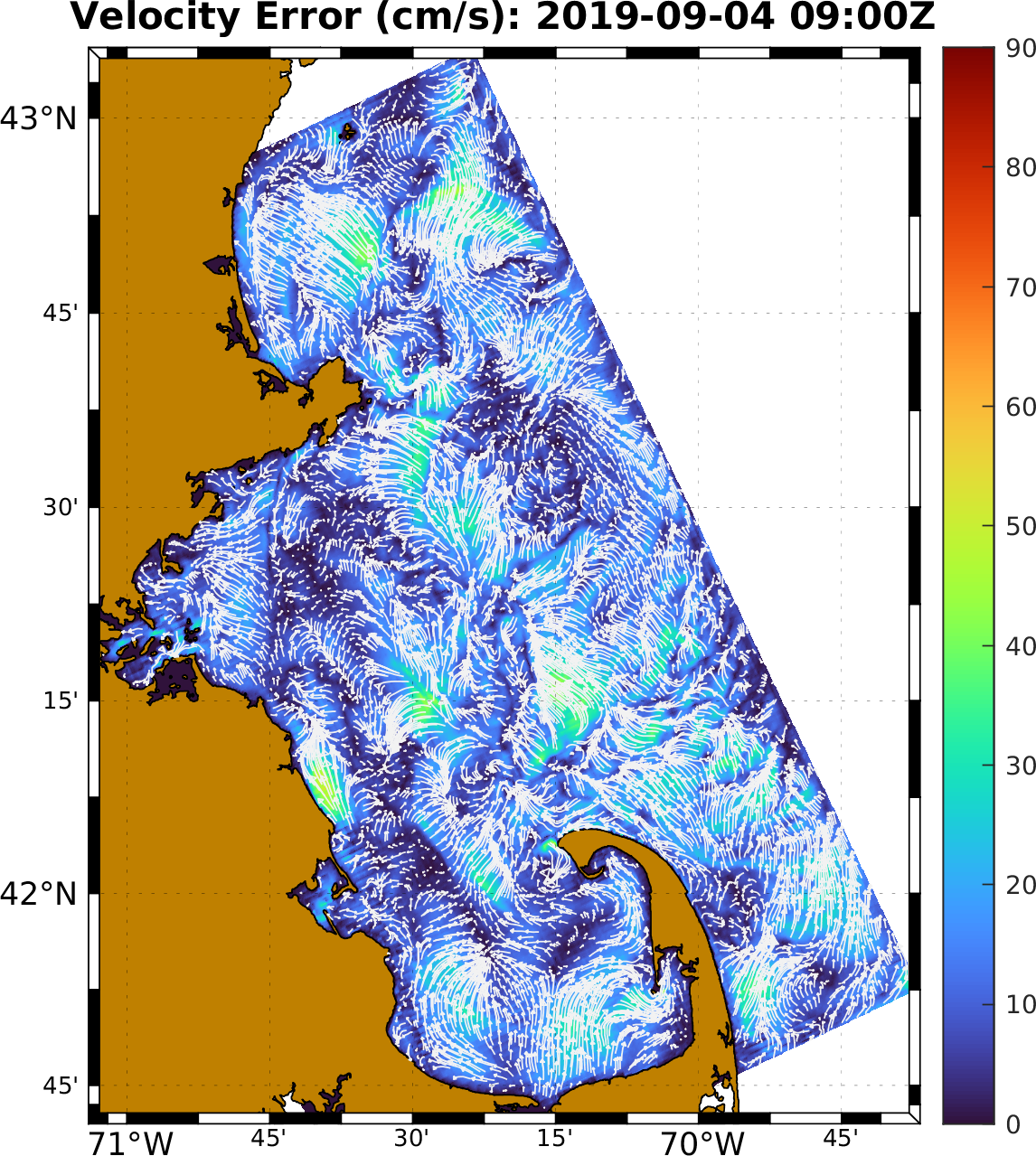}}
    \subfloat[]{\includegraphics[width=0.20\textwidth]{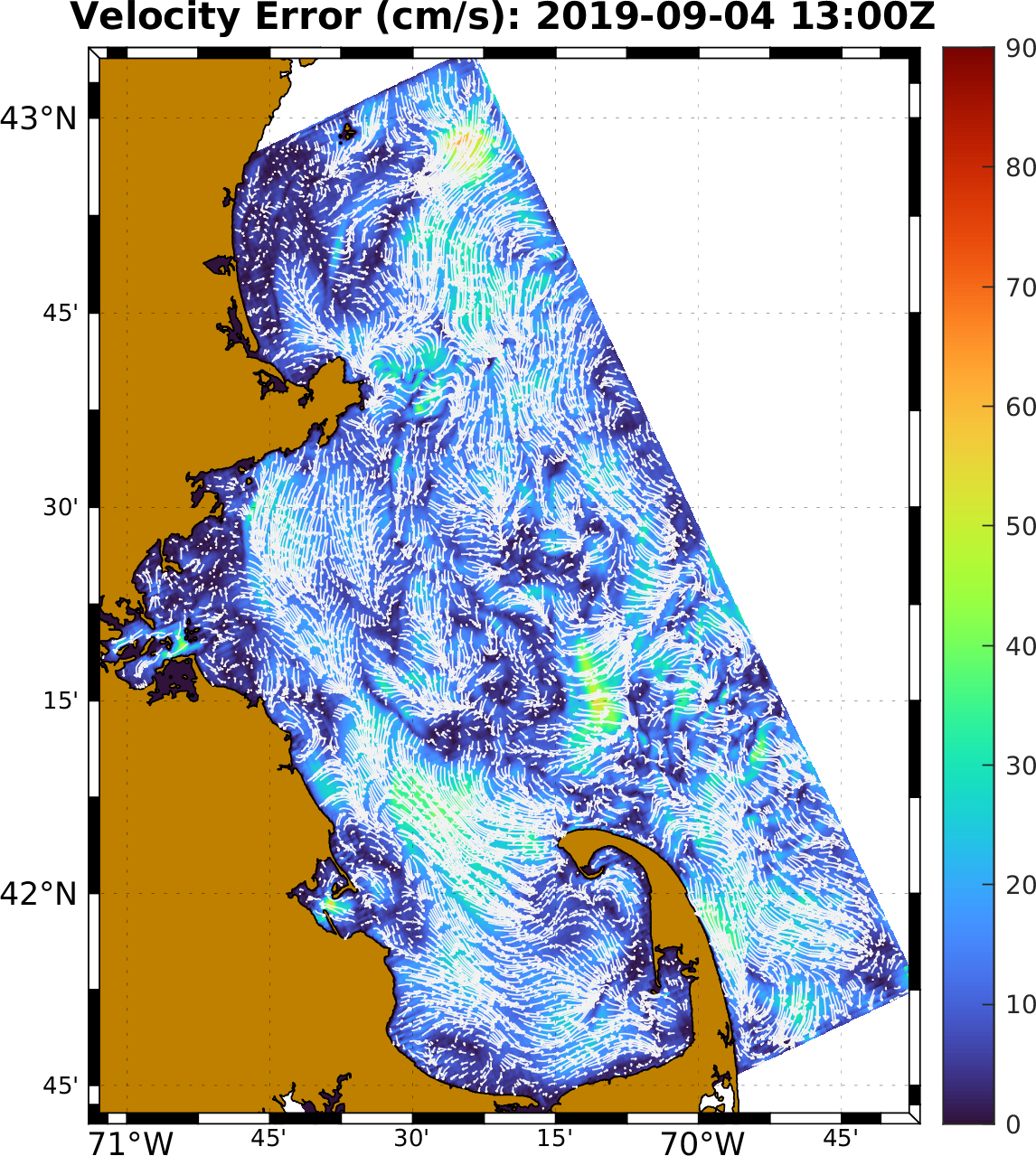}}
    \caption{\small \emph{Applying FCN in MB}. Surface velocity field amplitudes overlaid with curved velocity vectors. Ground truth from the MSEAS-PE forecast (top row), FCN predictions for training-run 264 (middle row), and error fields (bottom row) for five different forecast times: 2019 Sep 3 09Z and 6, 12, 24, and 28 hours later.}\label{fig:massbay_fields}
\end{figure*}

We now evaluate FCN based on a real ocean application in MB.
MB is a sub-region of the Gulf of Maine (GoM), off the northeast US coast.
The circulation in MB is commonly from north to south and is remotely driven from the GoM coastal current and mean wind stress \cite{blumberg_et_al_1993,geyer_ledwell_mwra1997,lermusiaux_JMS2001,haley_et_al_Oceans2020}. 
However, the MB circulation varies both seasonally and in response to wind events.
MB is also strongly impacted by tides, with intense internal waves radiating off Stellwagen Bank \cite{halpern1969observations,
lermusiaux_JMS2001,ali_et_al_Oceans2023}.

\subsubsection{Test Case Description}
The present MB simulation was completed as a part of
our ``Bayesian Intelligent Ocean Modeling and Acidification Prediction Systems'' (BIOMAPS) research \cite{lermusiaux_et_al_Oceans2019,haley_et_al_Oceans2020,gupta_lermusiaux_PO2023}.
The MSEAS-PE modeling system was configured in a stand-alone mode with 333~m resolution ($266\times451$ horizontal grid) 
and 100 vertical levels optimized for MB.
The sub-tidal initial and boundary conditions were downscaled from the global $1/12^\circ$ Hybrid Coordinate Ocean Model (HYCOM) analyses \cite{cummings_smedstad_2013}, using our optimization for our higher resolution coastlines and bathymetry \cite{haley_et_al_OM2015}. Local corrections were made using feature models and synoptic CTDs. Tidal forcing was computed from the OSU high-resolution TPXO8-Atlas, by reprocessing for our higher resolution bathymetry/coastline and quadratic bottom drag (a nonlinear extension of \cite{logutov_lermusiaux_OM2008}). The atmospheric forcing consisted of hourly analyses/forecasts of wind stresses, net heat flux, and surface freshwater flux from the 3\;km North American Mesoscale Forecast System (NAM) \cite{ncep_nam_2019}.
The MB reanalysis spanned the period August\;11 to September\;13, 2019, and was saved at hourly resolution, providing 781 snapshots.
Additional details can be found in \cite{haley_et_al_Oceans2020}.

The FCN results presented in this paper are for the surface currents. 
The first 48 snapshots are not used (dynamical adjustments) and 733 snapshots of
size 2x450x264 are available (each snapshot is an array for $u$ and $v$ in cm/s). 
The snapshots contain both land and ocean regions with 39,697 land cells and 79,103 ocean cells.

\emph{ML data usage, NN architecture, loss function, parameters, and sensitivity studies}.
For the results we show, we utilize the first $400$ snapshots for training, the next $100$ for validation, and the remaining 233 for verification/prediction.
We performed a sensitivity study of the FCN architectural choices and hyperparameters similar to that for MAB (Sect.\;\ref{sec:test_case_MAB}).
For the field results shown next, the
MB-trained FCN model corresponds to MB-training-run\;264 that has a tuned global batch size of 6, size of image patches of 3x3, number of AFNO layers (depth) of 10, and embedded channel size of 384.

\subsubsection{Learning Results}

A comparison of the simulated ground truth, MSEAS-PE forecasts of the surface velocity field for 2019 Sep 3-4 in MB, to FCN predictions 
trained on earlier MSEAS-PE fields (MB-training-run 264) is shown in Figure\;\ref{fig:massbay_fields}.
The circulation is complex with dynamic features and rapid changes.
Tides have a strong influence, especially over Stellwagen Bank and around the
tip and eastern side of Cape Cod. The winds are moderate and modulate the tidal effects. 
We note that winds were not contained in the FCN inputs so FCN can only learn and predict their effects from indirect surface current responses.
FCN does a good job capturing the tidal signal and the general sense of the local flows, 
but somewhat
underestimates the intensification over Stellwagen Bank.

\begin{figure}[h!]
    {\includegraphics[width=1\linewidth]{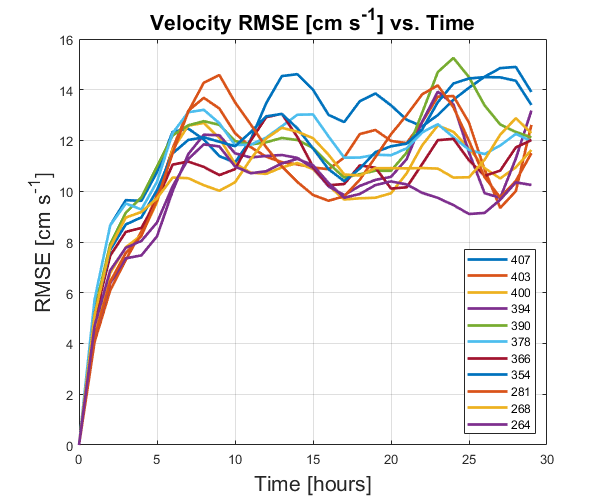}}
    \caption{\small \emph{Applying FCN in MB}. 
    Errors (RMSE) of FCN forecasts of surface velocity for 0 to 29h forecast lead times (2019 Sep 3 08Z to Sep 4 13Z). The FNC forecasts correspond to ten MAB training-runs with different hyperparameter values.
    }
    \label{fig:massbay_1D_overlay}
\end{figure}

In Figure\;\ref{fig:massbay_1D_overlay}, we show RMSEs of ten FCN forecasts of MB surface velocity fields for 0 to 29h forecast lead times. The different FCN forecasts were selected among the better ones (one of which is MB-training-run 264 shown on Fig.\;\ref{fig:massbay_fields}) and correspond to ten MAB training-runs with different hyperparameter values.
For these FCN forecasts,
averaged errors grow linearly quickly, and then taper off, with 
clear effects of tides and moderate winds.
Results confirm that the FCN forecasts have accuracy at many locations in MB, with most local errors remaining within 1 to 50\% of the variability after 29\;h. The FCN forecast uncertainty is estimated by the spread of the RMSE curves shown. 

\section{Conclusions}

Deep neural operator modeling frameworks have been recently developed for predicting the evolution of dynamical systems, learning from past simulation data or experimental data. The present work investigated the use of such
models for reproducing and predicting the 2D flow past a cylinder and realistic ocean dynamics, learning from past simulation data. The flow past a cylinder example confirmed that FCN and L-DoN can predict idealized flows with periodic, asymmetric vortex shedding.
We then illustrated the use of deep neural operator models to forecast ocean surface circulation in the Middle Atlantic Bight and in Massachusetts Bay, using training and verification data from our multi-resolution, data-assimilative MSEAS-PE simulations that we employed during real-time sea experiments in 2006 and 2019, respectively. 
We completed some sensitivity studies on the ML model data, architecture, and parameters. 
We compared sets of ML forecasts to the ground-truth MSEAS-PE simulations, and we found that the trained deep neural operator models predicted several of the features and showed some skill.

The results show promise for future ocean research and applications. This includes targeted repetitive forecasting, data assimilation, path planning, adaptive sampling, monitoring, sustainability, protection, risk management, and climate.

\section*{Acknowledgments}
We thank the members of our MSEAS group for useful discussions.
We also thank Marius Wiggert (Berkeley), Manan Doshi (MIT-MSEAS), and Claire Tomlin (Berkeley) for their initial suggestions. 
We are grateful to the Office of Naval Research for partial support under grant N00014-20-1-2023 (MURI ML-SCOPE) and 
Tech Candidate grant N00014-21-1-2831 (Compression and Assimilation for Resource-Limited Operations) 
to the Massachusetts Institute of Technology. 
We also thank the HYCOM team for their ocean fields, as well as UCAR and NCEP for their atmospheric forcing forecasts and reanalyses. 

%

\small
\bibliographystyle{ieeetr}
\bibliography{nonlinear_roms,mseas,refsdmd,refs_fno_don,refs_oceans,refs_fluids}

\end{document}